\documentclass[10pt,journal,compsoc]{IEEEtran}

\usepackage{multirow}
\usepackage{xspace}
\usepackage{textcomp}
\usepackage{gensymb}
\usepackage{amssymb}
\usepackage[table, dvipsnames]{xcolor}

\newcommand*{\eg}{e.g.\@\xspace}
\newcommand*{\ie}{i.e.\@\xspace}
\newcommand{\etal}{{et al}.\@ }

\makeatletter
\newcommand*{\etc}{%
    \@ifnextchar{.}%
        {etc}%
        {etc.\@\xspace}%
}

\newcommand{\norm}[1]{\left\lVert#1\right\rVert_{1}}

\ifCLASSOPTIONcompsoc
  \usepackage[nocompress]{cite}
\else
  \usepackage{cite}
\fi
\ifCLASSINFOpdf
   \usepackage[pdftex]{graphicx}
\else
\fi
\usepackage{amsmath}
\usepackage{array}
\begin{document}
%
\title{Learning Invariance from Generated Variance for Unsupervised Person Re-identification}
%
%
%
%

\author{Hao Chen,
        Yaohui Wang,
        Benoit Lagadec, 
        Antitza Dantcheva, 
        Francois Bremond

\IEEEcompsocitemizethanks{
\IEEEcompsocthanksitem H.~Chen, Y.~Wang, A.~Dantcheva and F.~Bremond are with Inria and Université Côte d'Azur, 2004 Route des Lucioles, 06902 Valbonne, France.
E-mail: \{hao.chen, yaohui.wang, antitza.dantcheva, francois.bremond\}@inria.fr
\IEEEcompsocthanksitem B.~Lagadec is with European Systems Integration, 362 Avenue du Campon, 06110 Le Cannet, France. E-mail: benoit.lagadec@esifrance.net} 
}

%
%

\markboth{}%
{Shell \MakeLowercase{\textit{et al.}}: Bare Advanced Demo of IEEEtran.cls for IEEE Computer Society Journals}
%



\IEEEtitleabstractindextext{%
\begin{abstract}
This work focuses on unsupervised representation learning in person re-identification (ReID). Recent self-supervised contrastive learning methods learn invariance by maximizing the representation similarity between two augmented views of a same image. However, traditional data augmentation may bring to the fore undesirable distortions on identity features, which is not always favorable in id-sensitive ReID tasks. In this paper, we propose to replace traditional data augmentation with a generative adversarial network (GAN) that is targeted to generate augmented views for contrastive learning. A 3D mesh guided person image generator is proposed to disentangle a person image into id-related and id-unrelated features. Deviating from previous GAN-based ReID methods that only work in id-unrelated space (pose and camera style), we conduct GAN-based augmentation on both id-unrelated and id-related features. We further propose specific contrastive losses to help our network learn invariance from id-unrelated and id-related augmentations. By jointly training the generative and the contrastive modules, our method achieves new state-of-the-art unsupervised person ReID performance on mainstream large-scale benchmarks. 
\end{abstract}

\begin{IEEEkeywords}
Person re-identification, image synthesis, representation disentanglement, data augmentation,  contrastive learning
\end{IEEEkeywords}}

\maketitle

\IEEEdisplaynontitleabstractindextext

%
\IEEEpeerreviewmaketitle

\ifCLASSOPTIONcompsoc
\IEEEraisesectionheading{\section{Introduction}\label{sec:introduction}}
\else
\section{Introduction}
\label{sec:introduction}
\fi

%
%
%
%

\IEEEPARstart{G}{iven} an image of a target person, a person re-identification (ReID) system~\cite{Ye_2021_reidsurvey, Karanam2019ASE} aims at matching images of the same person across non-overlapping cameras. With the help of human-annotated labels, \textit{supervised person ReID} methods~\cite{sun2018beyond,Chen_2020_WACV} have yielded impressive results. However, there usually exist strong domain gaps between different domains, such as illumination condition, camera property and scenario variation. As shown in previous methods~\cite{Song_2019_CVPR,Jin_2020_CVPR}, a ReID model trained on a specific domain is hard to generalize to other domains. One straightforward solution is to annotate and re-train the ReID model in a new domain, which is cumbersome and time-consuming for real-world deployments. Towards an automatic adaptive system,\textit{ unsupervised person ReID} ~\cite{zhong2019invariance,ge2020mutual,Chen_2021_joint} has attracted increasing attention in the research community. Compared with supervised counterparts, unsupervised methods directly learn from unlabeled images and therefore entail better scalability in real-world deployments.

Recent \textit{self-supervised contrastive learning} studies~\cite{chen2020simple,He_2020_CVPR} have shown promising performance in unsupervised representation learning. By maximizing the representation similarity between two different views (augmented versions) of a same image, contrastive methods learn representations that are invariant to different conditions. In this context, data augmentation plays a crucial role in mimicking real-world condition variance. Contrastive learning methods are able to build more robust representations, given they were provided with better augmented views. Previous methods generally consider traditional data augmentation techniques, \eg, random flipping, cropping, color jittering, blurring and erasing~\cite{Zhong2020RandomED}. However, these random augmentation techniques may cause undesirable distortion to crucial identity information. To overcome this issue, we propose to use a Generative Adversarial Network (GAN)~\cite{goodfellow2014generative} as an augmentation substitute, as it is able to disentangle a representation into id-related and id-unrelated features (see Table~\ref{Tab: Id-related and Id-unrelated}). More accurate augmented views can be obtained by modifying a certain factor while preserving other factors.

Previous GAN-based unsupervised ReID methods \cite{wei2018person,bak2018domain,Zhong_2018_ECCV,Zou2020JointDA} often treat unsupervised ReID as an unsupervised domain adaptation task, which attempts to adapt a model trained on a labeled source domain to an unlabeled target domain. Under this setting, it is intuitive to use GAN-based style transfer~\cite{huang2017arbitrary, isola2017image} to generate source domain images in the style of a target domain. A model can be re-trained on the generated images in target domain style with source domain labels. However, unsupervised domain adaptation performance often strongly relies on quality and scale of the source domain. Differently, we treat unsupervised ReID as a \textit{contrastive representation learning} task, where the source domain is not mandatory. To this end, we integrate a generative module and a contrastive module into a joint learning framework. 

For the generative module, we propose a 3D mesh based generator. Conventional pose transfer methods~\cite{ge2018fd,li2019cross} use 2D pose~\cite{cao2017realtime} to guide generation, not preserving body shape information. 3D mesh recovery~\cite{kanazawaHMR18} jointly estimates body shape, as well as 3D pose, which conserves more identity information for unsupervised ReID. We use 3D meshes to guide the generation, where generated images in new poses are then used as augmented views in the contrastive module.

For the contrastive module, we use a clustering algorithm to generate pseudo labels, aimed at maximizing representation similarity between different views of a same pseudo identity. Our model attracts a generated view to its original view, while repulsing the generated view from images of different identities. The contrastive module permits an identity encoder to extract view-invariant identity features, which, in turn, improves the generation quality.

\begin{table}[t]
\caption{Id-related and Id-unrelated factors in a person image.}
\label{Tab: Id-related and Id-unrelated}
\centering
\begin{tabular}{c|c}
\hline
Id-related  & Id-unrelated \\
\hline
cloth color,  & pose, view-point, \\
hair color, texture, & illumination, camera style\\
body shape & background\\
\hline
\end{tabular}
\end{table}

In our previous work~\cite{Chen_2021_joint}, GAN-based augmentation was only conducted on id-unrelated features, which has been common practice in previous GAN-based ReID methods~\cite{Zhong2018CameraSA,ge2018fd,zheng2019joint}. Modifying id-unrelated features allows for learning identity features that are more invariant to id-unrelated variations. In this paper, we explore the possibility of conducting GAN-based augmentation on the id-related features to further improve the ReID performance. Inspired by Mixup~\cite{zhang2018mixup} that interpolates two images to learn a smoother decision boundary between two classes, we propose to interpolate disentangled id-related features inside the generative module, namely \textbf{Disentangled Mixup (D-Mixup)}. 
As shown in Table 2, if two persons $P_1$ and $P_2$ respectively wear red and yellow clothes, an in-between identity in orange clothes should be marked as $0.5P_1+0.5P_2$. However, in a dataset, such a person in orange clothes is normally labeled as a totally different identity $P_3$, which hinders a network from learning the accurate relationship between different identities. Compared to traditional image-level Mixup~\cite{zhang2018mixup} and feature-level Mixup~\cite{Verma2019ManifoldMB}, our proposed D-Mixup generates more accurate in-between identity images, which are more suitable for fine-grained person ReID. In our D-Mixup, we try to make our network understand the mixed identity $0.5P_1+0.5P_2$ is not related to id-unrelated features (pose and view-point), but only related to id-related features (cloth color). 

To summarize, our contributions include the following:
\begin{itemize}
  \item We propose a 3D mesh guided generator to disentangle representations into id-related and id-unrelated features. Two novel data augmentation techniques are proposed respectively on id-unrelated and id-related features.
  \item We propose Rotation Contrast and Mixup Contrast modules to respectively learn invariance from id-unrelated and id-related augmented views.
  \item We propose an enhanced joint generative and contrastive learning framework. We comprehensively investigate how the generative and contrastive modules mutually promote each other and contribute to unsupervised ReID performance.
  \item Extensive experiments validate the superiority of proposed GAN-based augmentation over traditional augmentation for unsupervised person ReID. Our method achieves new state-of-the-art unsupervised person ReID performance on mainstream image-based datasets, including Market-1501, DukeMTMC-reID and MSMT17. 
  \item Our method can be also applied to video-based person ReID. Our method significantly outperforms previous unsupervised video person ReID methods on MARS and DukeMTMC-VideoReID datasets.
\end{itemize}

\setlength{\tabcolsep}{4pt}

\begin{table}[t]
\caption{Interpolation results between two random persons $P_1$ and $P_2$ with image-level Mixup~\cite{zhang2018mixup}, feature-level Mixup (F-Mixup)~\cite{Verma2019ManifoldMB} and our proposed disentangled Mixup (D-Mixup). To visualize results from F-Mixup, we follow AMR~\cite{beckham2019adversarial} to train a VAE-GAN for mixed image reconstruction. Our D-Mixup only interpolates disentangled identity features in the generation, which alleviates noise from mixed structural features.}
\label{Tab: Mixup comparison}
\centering
\begin{tabular}{c|cc|c|c|cc}
  & \multicolumn{2}{c|}{Inputs} & Mixup & F-Mixup& \multicolumn{2}{c}{D-Mixup} \\\hline

Image & \raisebox{-.4\height}{\includegraphics[scale=0.2]{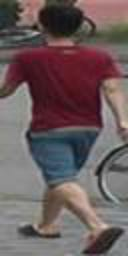}}  & \raisebox{-.4\height}{\includegraphics[scale=0.2]{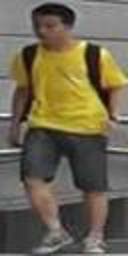}} & \raisebox{-.4\height}{\includegraphics[scale=0.2]{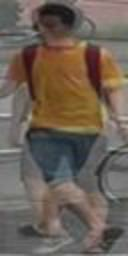}} &
\raisebox{-.4\height}{\includegraphics[scale=0.2]{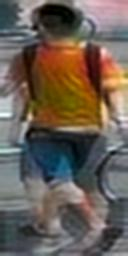}} &
\raisebox{-.4\height}{\includegraphics[scale=0.2]{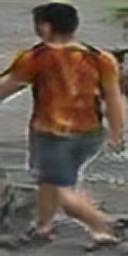}}&
\raisebox{-.4\height}{\includegraphics[scale=0.2]{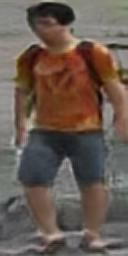}}\\
\hline
Image & \raisebox{-.4\height}{\includegraphics[scale=0.2]{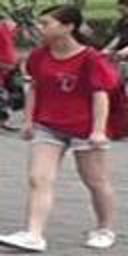}}  & \raisebox{-.4\height}{\includegraphics[scale=0.2]{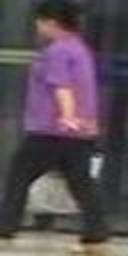}} & \raisebox{-.4\height}{\includegraphics[scale=0.2]{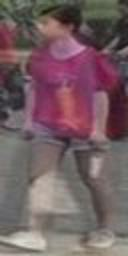}} &
\raisebox{-.4\height}{\includegraphics[scale=0.2]{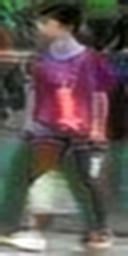}}&
\raisebox{-.4\height}{\includegraphics[scale=0.2]{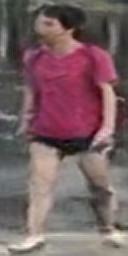}}&
\raisebox{-.4\height}{\includegraphics[scale=0.2]{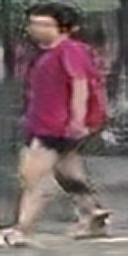}}\\
\hline
\multirow{2}{*}{Label}  & $1.0P_1$ & $0.0P_1$ & $0.5P_1$ & $0.5P_1$ & $0.5P_1$ & $0.5P_1$ \\
  & $0.0P_2$ & $1.0P_2$ & $0.5P_2$ & $0.5P_2$ & $0.5P_2$ & $0.5P_2$ \\
\hline
\end{tabular}
\end{table}

\setlength{\tabcolsep}{6pt}

 




\section{Related Work}
\subsection{Contrastive learning}
Contrastive learning~\cite{hadsell2006dimension} has shown impressive performance for un-/self-supervised representation learning~\cite{Wu2018UnsupervisedFL,chen2020simple,He_2020_CVPR,caron2020unsupervised,grill2020bootstrap,chen2021exploring}. Such contrastive methods target at learning representations that are invariant to different distortions by attracting positive pairs, while repulsing negative pairs. For each image, a positive pair can be constituted by two augmented views, whereas all other images in a dataset are regarded as negative samples. Contrastive learning methods benefit from a set of well defined data augmentation techniques, which can mimic real-world image distortions. For example, MoCo~\cite{He_2020_CVPR} used random cropping, color jitterring, horizontal flipping and grayscale conversion to obtain positive view pairs. As an extension, MoCo-v2~\cite{chen2020improved} included blurring and stronger color distorsion, which enhanced the original method. However, most of data augmentation settings in contrastive learning methods were designed for general image classification datasets, \eg, ImageNet~\cite{Russakovsky2015ImageNetLS}. These traditional augmentation techniques are not always suitable for color-sensitive person ReID, especially those that introduce strong color distorsion.

\subsection{Data augmentation}
As a technique to constitute positive pairs, data augmentation plays an important role in contrastive learning. Recently, GAN and Mixup have provided new approaches for data augmentation in person ReID.

\subsubsection{GAN-based augmentaion}
Zheng \etal~\cite{zheng2017unlabeled} unconditionally generated a lot of unlabeled person images with DCGAN~\cite{Radford2016UnsupervisedRL} to enlarge data volume for supervised ReID. Following GAN-based methods were usually conditionally conducted on some factors from Table~\ref{Tab: Id-related and Id-unrelated}. \textbf{1) Pose:} With the guidance of 2D poses, FD-GAN~\cite{ge2018fd} and PN-GAN~\cite{qian2018pose} generated a target person in new poses to learn pose-irrelevant representations for single-domain supervised ReID. Similar pose transfer \cite{li2019cross} was then proposed to address unsupervised domain adaptive (UDA) ReID. \textbf{2) Dataset style (illumination):} As a dataset is usually recorded in a uniform illumination condition, PTGAN~\cite{wei2018person} and SyRI~\cite{bak2018domain} used CycleGAN~\cite{zhu2017unpaired} to minimize the domain gap between different datasets by generating person images in the style of a target domain. \textbf{3) Camera style:} Instead of the general dataset style, CamStyle~\cite{Zhong2018CameraSA} transferred images captured from one camera into the style of another camera, in order to reduce inter-camera style gaps. Similar method~\cite{Zhong_2018_ECCV} was then applied to UDA ReID. \textbf{4) Background:} SBSGAN~\cite{Huang_2019_ICCV} and CR-GAN~\cite{chen2019instance} respectively were targeted at removing and switching the background of a person image to mitigate background influence for UDA ReID. \textbf{5) General structure:} By switching global and local level identity-unrelated features, IS-GAN~\cite{NIPS2019_8771} disentangled a representation into identity-related and identity-unrelated features without any concrete guidance. As a concrete guidance, a gray-scaled image contains multiple id-unrelated factors of a person image, including pose, background and carrying structures. By recoloring gray-scaled person images with the color distribution of other images, DG-Net~\cite{zheng2019joint} and DG-Net++~\cite{Zou2020JointDA} learned disentangled identity representations invariant to structure factors. Our proposed 3D mesh guided generator shares certain similarity with pose transfers and DG-Net++. However, both pose transfers and DG-Net++ lose body shape information, which can be conserved by 3D meshes. Moreover, as opposed to DG-Net++, we do not transfer style in a cross-domain manner, which allows our method to operate without a source domain. 

\subsubsection{Mixup}
Mixup~\cite{zhang2018mixup} is a simple yet effective data augmentation technique that interpolates two samples and labels into one new in-between sample, which encourages a smoother decision boundary between two classes. The interpolation can be conducted between two images ~\cite{zhang2018mixup,Tokozume2018BetweenClassLF}, two feature representations~\cite{Verma2019ManifoldMB} and two portions of different images~\cite{Yun2019CutMixRS}. Initially proposed for supervised image classification~\cite{zhang2018mixup,Tokozume2018BetweenClassLF}, Mixup has been successfully extended to semi-supervised learning~\cite{berthelot2019mixmatch,Berthelot2020ReMixMatch:}, unsupervised domain adaptation~\cite{Xu2020AdversarialMixup}, as well as novel class discovery~\cite{zhong2021openmix}. AugMix~\cite{Hendrycks2020AugMixAS} combines multiple augmented versions of an image into a mixed image and proves that such technique can enhance robustness on corrupted data. CAIL~\cite{luo2020generalizing} applies image-level Mixup between a source domain image and a target domain image to create a between-domain person image, which facilitates cross-domain knowledge transfer in unsupervised domain adaptive ReID. 
The above methods usually interpolate whole images or whole representations, resulting in noise from overlapping person structures. To reduce noise from mixed person structures, we propose to interpolate only disentangled identity features, which is compatible with our proposed 3D mesh guided GAN.

\subsection{Unsupervised person ReID}
Depending on the necessity of a large-scale labeled source dataset, unsupervised person ReID methods can be roughly categorized into unsupervised domain adaptive (UDA) and fully unsupervised ReID. We note that above mentioned GAN-based unsupervised ReID  methods~\cite{li2019cross,wei2018person,bak2018domain,Zhong_2018_ECCV,chen2019instance,Zou2020JointDA} fall into the setting of UDA ReID. Several works~\cite{Wang2018TransferableJA,Lin2018MultitaskMF} leveraged semantic attributes to facilitate the domain adaptation. Another prominent approach has to do with assigning pseudo labels to unlabeled images and conducting pseudo label learning~\cite{Yu2019UnsupervisedPR,fu2019self, yang2020asymmetric, ge2020mutual, luo2020generalizing, zhong2019invariance, zhong2020learning}. Pseudo labels can be obtained by existing clustering algorithms, \eg, K-means~\cite{ge2020mutual} and DBSCAN~\cite{Zou2020JointDA,yang2020asymmetric}, or newly designed pseudo labelling algorithms~\cite{Yu2019UnsupervisedPR,zhong2020learning}. Since the performance of UDA ReID is highly correlated to the scale and quality of a source domain, recent fully unsupervised ReID methods have attracted more attention. Most of previous fully unsupervised methods~\cite{Lin2019ABC,Lin2020UnsupervisedPR,Wang_2020_CVPR,li2020joint, wu2020tracklet} were based on pure pseudo label learning. Our previous method GCL~\cite{Chen_2021_joint} has entailed a hybrid GAN and pseudo label learning method, which is compatible with both UDA and fully unsupervised settings. We here propose a new id-related augmentation D-Mixup, which enhances our framework to achieve new state-of-the-art performance under both UDA and fully unsupervised settings.  

\begin{figure*}
\centering
   \includegraphics[width=0.97\linewidth]{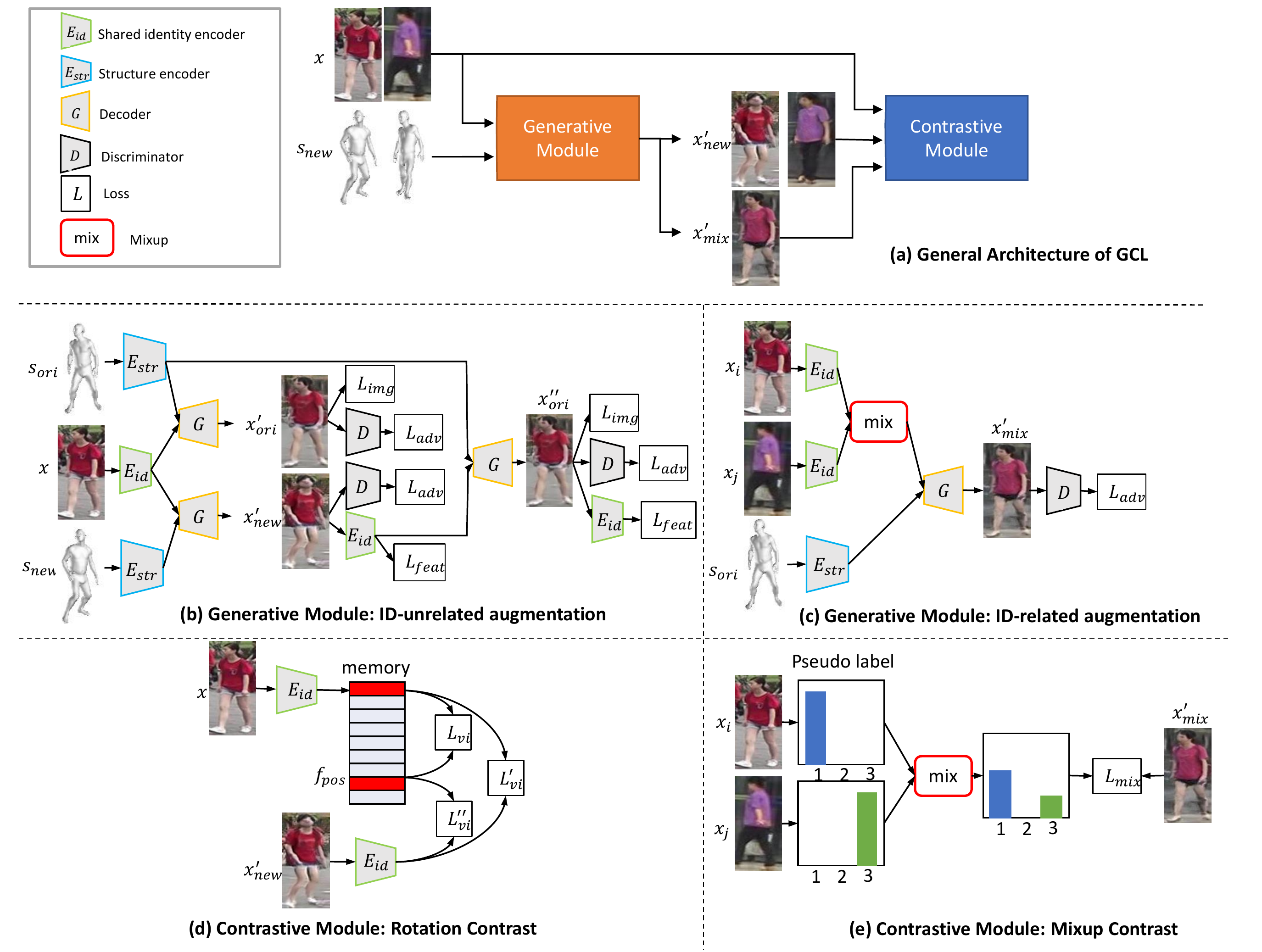}
   \caption{\textbf{(a) General architecture of GCL+}: The framework is composed of a generative module \textbf{(b, c)} and a contrastive module\textbf{ (d, e)}, which are coupled by the shared identity encoder $E_{id}$. \textbf{(b) Mesh rotation (id-unrelated augmentation) }: The decoder $G$ combines the identity features encoded by $E_{id}$ and structure features $E_{str}$ to generate an augmented view $x_{new}^{\prime}$ with a cycle consistency. \textbf{(c) D-mixup (id-related augmentation)}: The decoder $G$ generates a identity-mixed augmented view $x_{mix}^{\prime}$ with the mixed identity features. \textbf{(d) Rotation Contrast}: Viewpoint-invariance is enhanced by maximizing the agreement between original $E_{id}(x)$, synthesized $E_{id}(x_{new}^{\prime})$ and memory $f_{pos}$ representations. \textbf{(e) Mixup Contrast}: A smoother decision boundary can be learnt with $x_{mix}^{\prime}$ and the interpolated pseudo label. }
\label{fig:general structure}
\end{figure*}

\section{Method}
In this paper, we propose an enhanced joint \textbf{G}enerative and \textbf{C}ontrastive \textbf{L}earning (\textbf{GCL+}) for unsupervised person ReID. We define unsupervised ReID as a problem of learning invariance from self-augmented variance. As illustrated in Fig.~\ref{fig:general structure}.~(a), the proposed GCL+ constitutes of two modules: a generative module that provides GAN-based augmented views, as well as a contrastive module that learns invariance from augmented views. These two modules are coupled by a shared identity encoder. After the joint training, only the shared identity encoder is conserved for inference. In the following sections, we proceed to provide details related to both modules. To facilitate the reading, we include a list of abbreviations in Supplementary Materials Section C.

\subsection{Generative Module}
Our generative module is composed of 4 networks, including an identity encoder $E_{id}$, a structure encoder $E_{str}$, a decoder $G$ and a discriminator $D$. Given an unlabeled person ReID dataset $\mathcal{X}=\{x_1, x_2, ..., x_N\}$, we use the prominent algorithm HMR~\cite{kanazawaHMR18} to generate corresponding 3D meshes, which are then used as structure guidance in the generative module. By recoloring a specific 3D mesh to reconstruct a real image, a person representation can be disentangled into identity and structure features. We conduct data augmentation in two pathways: one on id-unrelated structure features with rotated meshes, the other one on identity features with D-Mixup.

\begin{table}[t]
\caption{Examples of 3D mesh guided generation on Market-1501 dataset. Each mesh is rotated by $45^{\degree}, 90^{\degree}, 135^{\degree}, 180^{\degree}, 225^{\degree}, 270^{\degree}$ and $315^{\degree}$. }
\label{Tab: Rotation}
\centering
\setlength\tabcolsep{2pt}
\begin{tabular}{ccccccccc}
  0\degree& &45\degree & 90\degree & 135\degree & 180\degree & 225\degree& 270\degree& 315\degree \\
\raisebox{-.4\height}{\includegraphics[scale=0.4]{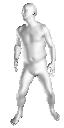}} &
$\to$&
\raisebox{-.4\height}{\includegraphics[scale=0.4]{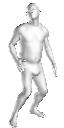}} & \raisebox{-.4\height}{\includegraphics[scale=0.4]{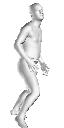}}& \raisebox{-.4\height}{\includegraphics[scale=0.4]{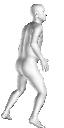}}& \raisebox{-.4\height}{\includegraphics[scale=0.4]{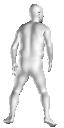}}& \raisebox{-.4\height}{\includegraphics[scale=0.4]{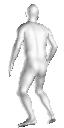}} &
\raisebox{-.4\height}{\includegraphics[scale=0.4]{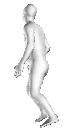}} &
\raisebox{-.4\height}{\includegraphics[scale=0.4]{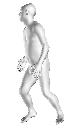}}
\\  
\raisebox{-.4\height}{\includegraphics[scale=0.2]{IMAGE/mixup2/0011_c3s3_075919_03_10.png}} &
$\to$&
\raisebox{-.4\height}{\includegraphics[scale=0.2]{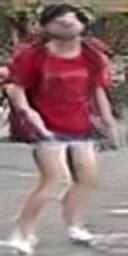}} & \raisebox{-.4\height}{\includegraphics[scale=0.2]{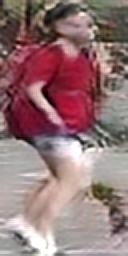}}& \raisebox{-.4\height}{\includegraphics[scale=0.2]{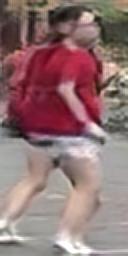}}& \raisebox{-.4\height}{\includegraphics[scale=0.2]{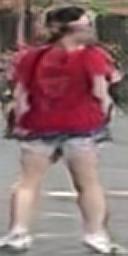}}& \raisebox{-.4\height}{\includegraphics[scale=0.2]{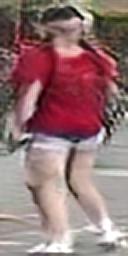}} &
\raisebox{-.4\height}{\includegraphics[scale=0.2]{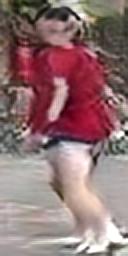}} &
\raisebox{-.4\height}{\includegraphics[scale=0.2]{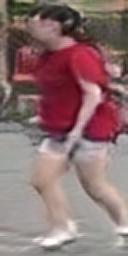}}
\\
\raisebox{-.4\height}{\includegraphics[scale=0.4]{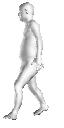}} &
$\to$&
\raisebox{-.4\height}{\includegraphics[scale=0.4]{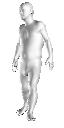}} & \raisebox{-.4\height}{\includegraphics[scale=0.4]{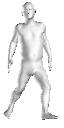}}& \raisebox{-.4\height}{\includegraphics[scale=0.4]{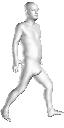}}& \raisebox{-.4\height}{\includegraphics[scale=0.4]{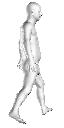}}& \raisebox{-.4\height}{\includegraphics[scale=0.4]{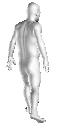}} &
\raisebox{-.4\height}{\includegraphics[scale=0.4]{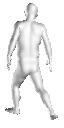}} &
\raisebox{-.4\height}{\includegraphics[scale=0.4]{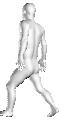}}
\\
\raisebox{-.4\height}{\includegraphics[scale=0.2]{IMAGE/mixup2/0093_c6s1_015001_02_10.png}} &
$\to$&
\raisebox{-.4\height}{\includegraphics[scale=0.2]{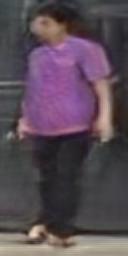}} & \raisebox{-.4\height}{\includegraphics[scale=0.2]{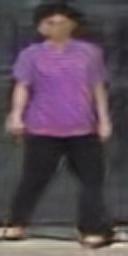}}& \raisebox{-.4\height}{\includegraphics[scale=0.2]{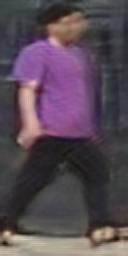}}& \raisebox{-.4\height}{\includegraphics[scale=0.2]{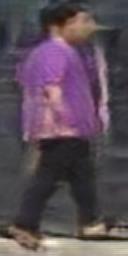}}& \raisebox{-.4\height}{\includegraphics[scale=0.2]{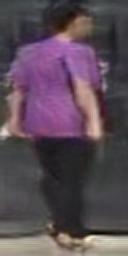}} &
\raisebox{-.4\height}{\includegraphics[scale=0.2]{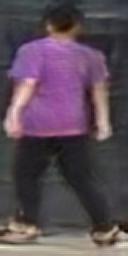}} &
\raisebox{-.4\height}{\includegraphics[scale=0.2]{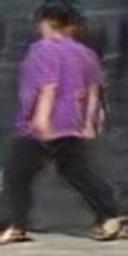}}\\
\end{tabular}
\end{table}

\subsubsection{Mesh-guided Rotation (id-unrelated augmentation)}
As shown in Fig.~\ref{fig:general structure}.~(b), given a person image and an estimated 3D mesh, we denote the 2D projection of the mesh as original structure $s_{ori}$. To mimic real-world camera view-point, as shown in Table~\ref{Tab: Rotation}, we rotate the 3D mesh by $45^{\degree}, 90^{\degree}, 135^{\degree}, 180^{\degree}, 225^{\degree}, 270^{\degree}$ and $315^{\degree}$ and randomly take one 2D projection from these rotated meshes as a new structure $s_{new}$. The unlabeled image is encoded to identity features by the identity encoder $E_{id}: x \to f_{id}$, while both original and new structures are encoded to structure features by the structure encoder $E_{str}: s_{ori} \to f_{str(ori)}, s_{new} \to f_{str(new)}$. Combining both identity and structure features, the decoder generates synthesized images $G: (f_{id}, f_{str(ori)}) \to x_{ori}^{\prime}, (f_{id}, f_{str(new)}) \to x_{new}^{\prime},$ where a prime is used to represent generated images. 

As we do not have real images in new structures (paired data), a cycle consistency reconstruction~\cite{zhu2017unpaired} becomes indispensable for the generative module. We encode the generated image in the new structure $x_{new}^{\prime}$ and decode once again to get synthesized images in original structures $G(E_{id}(x_{new}^{\prime}), s_{ori}) \to x_{ori}^{\prime\prime}$, where double primes denote cycle-generated images. We calculate a $\ell_1$ image reconstruction loss between the original image $x$, the generated image $x_{ori}^{\prime}$ and the cycle-generated image:
\begin{equation}
\begin{split}
\mathcal{L}_{img} = \mathop{\mathbb{E}} [\norm{x-x_{ori}^{\prime}}] + \mathop{\mathbb{E}} [\norm{x-x_{ori}^{\prime\prime}} ].
\end{split}
\label{image_recon_loss}
\end{equation}
To enhance the disentanglement in the cycle consistency reconstruction, we also calculate a $\ell_1$ feature reconstruction loss: 
\begin{equation}
\begin{split}
\mathcal{L}_{feat} = &\mathop{\mathbb{E}} [\norm{f_{id}-E_{id}(x_{new}^{\prime})}]+\\
&\mathop{\mathbb{E}} [\norm{f_{id}-E_{id}(x_{ori}^{\prime\prime})} ].
\end{split}
\label{feat_recon_loss}
\end{equation}

The discriminator $D$ attempts to distinguish between real and generated images with adversarial losses:
\begin{equation}
\begin{split}
\mathcal{L}_{adv} = &\mathop{\mathbb{E}} [ \log D(x) + \log (1-D(x_{ori}^{\prime})) ] + \\
&\mathop{\mathbb{E}} [ \log D(x) + \log (1-D(x_{new}^{\prime})) ]  + \\
&\mathop{\mathbb{E}} [ \log D(x) + \log (1-D(x_{ori}^{\prime\prime})) ].
\end{split}
\label{adv_loss}
\end{equation}

\begin{figure}[t]
\centering
\includegraphics[width=1\linewidth]{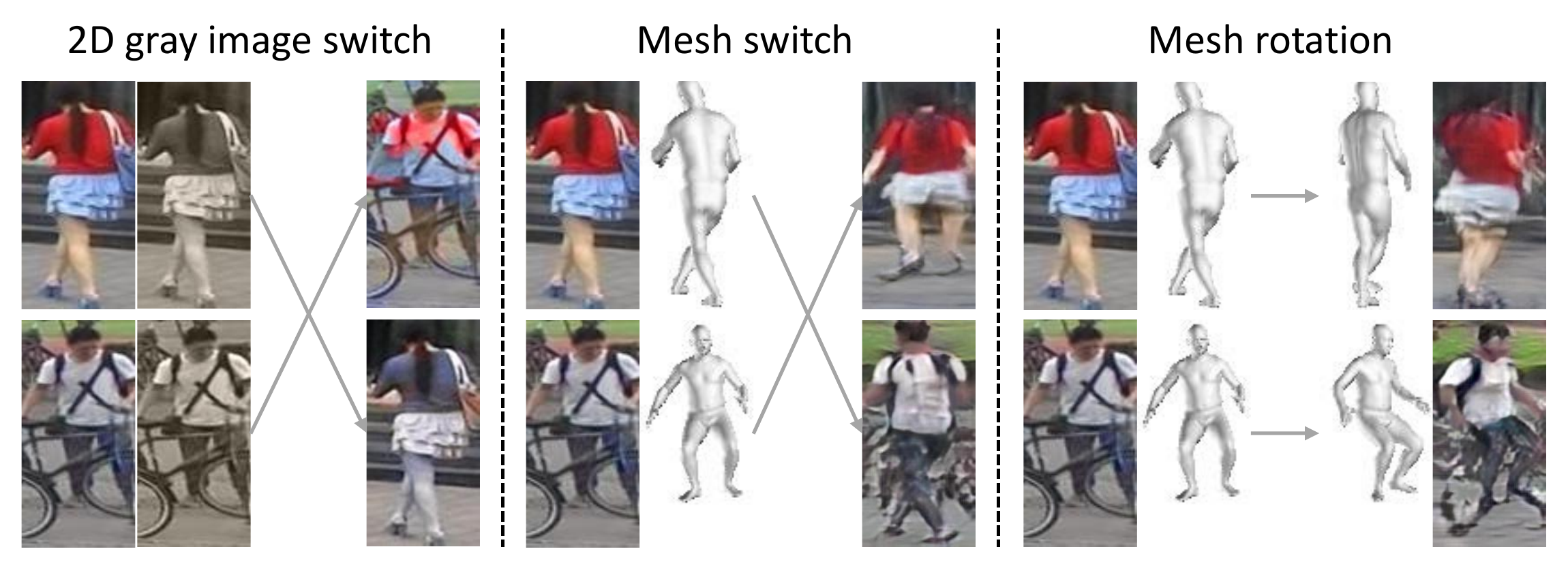}
\caption{Different ways of introducing structural variance (2D gray image switch~\cite{zheng2019joint}, Mesh switch and Mesh rotation) into generation.}
\label{fig_mesh_switch_mesh_rotation}
\end{figure}

\textbf{Remark.} As shown in Fig.~\ref{fig_mesh_switch_mesh_rotation}, we can switch 2D gray images~\cite{zheng2019joint,Zou2020JointDA}, switch meshes between random persons or rotate one's own mesh to introduce new structures as generation guidance. Although stronger pose and view-point variances can be introduced into generation, random switching hinders conservation of body shape information. After testing, we find that the most appropriate way to preserve body shape and generate accurate images is Mesh rotation, which yields higher performance in Table~\ref{table:mesh rotation mesh switch}.

\subsubsection{D-mixup (id-related augmentation)}\label{sec:D-mixup (id-related augmentation)}
\begin{figure}[t]
\centering
\includegraphics[width=0.9\linewidth]{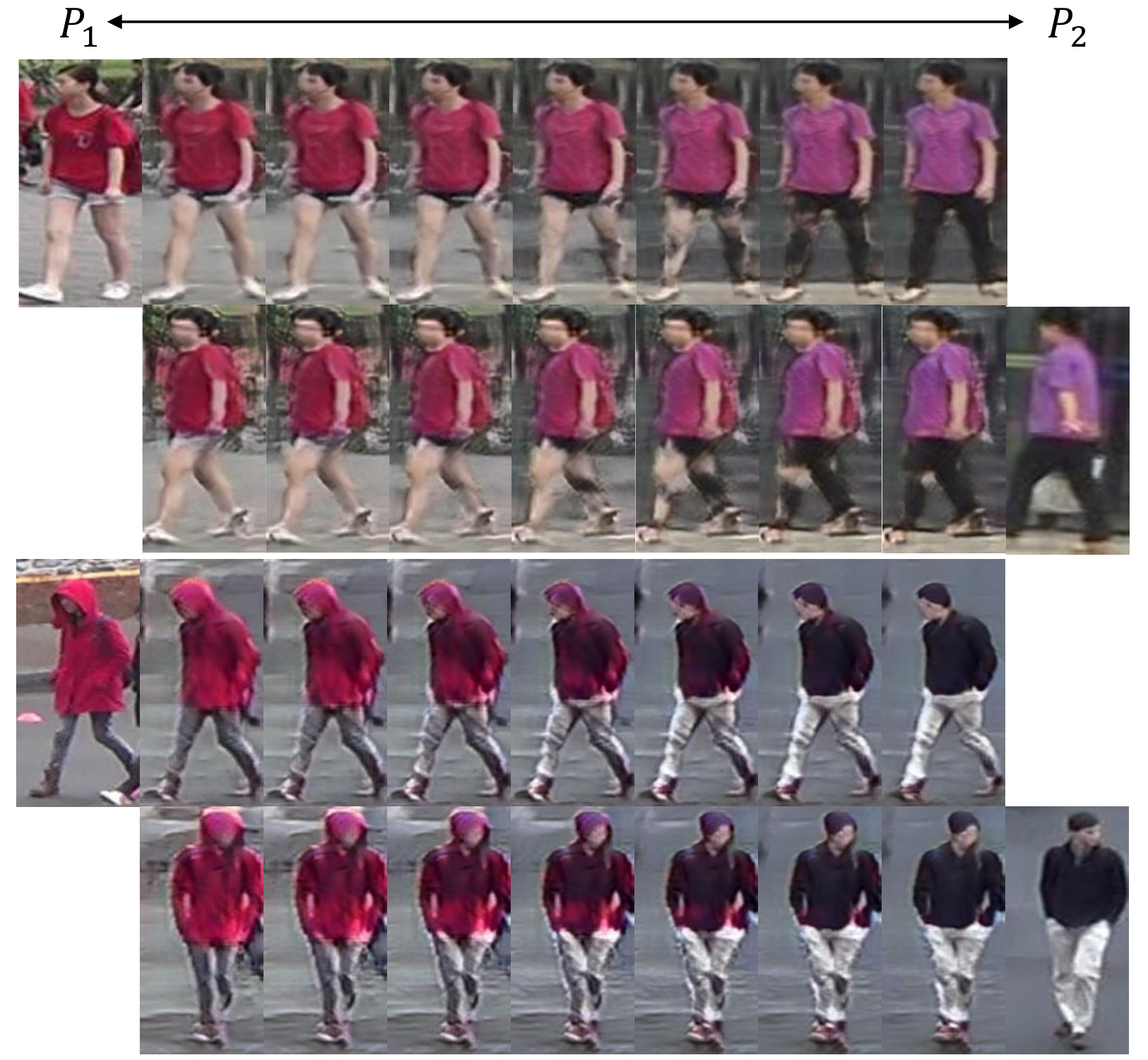}
\caption{Linear interpolation of disentangled identity features between two persons respectively from Market-1501 and DukeMTMC-reID.}
\label{fig_picture3_d_mixup}
\end{figure}

As shown in Fig.~\ref{fig:general structure}.~(c), given two random person images $x_{i}$ and $x_{j}$ in a mini-batch, we encode the images into identity features $E_{id}(x_{i}) \to f_{id(i)}$ and $E_{id}(x_{j}) \to f_{id(j)}$. We follow the original Mixup~\cite{zhang2018mixup} in using a Beta distribution with a hyper-parameter $\alpha$ to randomly sample a mixing coefficient $\lambda$:
\begin{equation}
\begin{split}
&\lambda = Beta(\alpha, \alpha),\ \lambda^* = max(\lambda, 1-\lambda)\\
&f_{id(mix)} = \lambda^* \cdot f_{id(i)} + (1-\lambda^*)\cdot f_{id(j)},
\end{split}
\label{mixup_feat}
\end{equation}
where $\lambda^*$ renders the mixed identity more similar to $x_{i}$. To conserve corresponding body shape information, we use the original structure of $x_{i}$, rather than $x_{j}$ as the generation guidance.
A mixed person image (see more interpolated examples in Fig.~\ref{fig_picture3_d_mixup}) can be generated by combining mixed identity features and original structure features $G(f_{id(mix)}, s_{ori(i)}) \to x_{mix}^{\prime}$. 
The discriminator $D$ attempts to distinguish between real and mixed images with the adversarial loss:
\begin{equation}
\begin{split}
\mathcal{L}_{adv\_mix} = &\mathop{\mathbb{E}} [ \log D(x) + \log (1-D(x_{mix}^{\prime})) ].
\end{split}
\label{mixup_adv_loss}
\end{equation}

More discussion about feature regularization losses is provided in Supplementary Materials Section A.

\subsubsection{Overall generative loss}
The overall GAN loss combines the above losses (\ref{image_recon_loss}), (\ref{feat_recon_loss}), (\ref{adv_loss}) and (\ref{mixup_adv_loss}) with a weighting coefficient $\lambda_{recon}$:
\begin{equation}
\begin{split}
\mathcal{L}_{gan} = \lambda_{recon}(\mathcal{L}_{img} +\mathcal{L}_{feat})+
   \mathcal{L}_{adv}+
   \mathcal{L}_{adv\_mix}.
\end{split}
\label{overall_gan_loss}
\end{equation}


\subsection{Contrastive Module}
The described generative module generates augmented views of a person image, which can form positive view pairs for the contrastive module. By maximizing similarity between positive pairs, the shared identity encoder is aimed at building robust representations that are invariant to distortions. For one identity, there are commonly several positive images in the dataset, which are recorded in different poses, camera styles and backgrounds. Only maximizing similarity between an image and its self-augmented views leads to sub-optimal performance. Moreover, previous methods~\cite{chen2020simple, He_2020_CVPR} have demonstrated the effectiveness of mining a large number of negative samples in contrastive learning. 

In order to mine more positives and a large number of negatives, we generate pseudo labels on a memory bank~\cite{Wu2018UnsupervisedFL} that stores all representations $\mathcal{M}$ corresponding to dataset images $\mathcal{X}$. Given a representation $f^t$ in the current epoch, the corresponding memory bank representation $\mathcal{M}[i]$ is updated with a momentum hyper-parameter $\beta$:
\begin{equation}
\mathcal{M}[i]^t =\beta\cdot \mathcal{M}[i]^{t-1} + (1-\beta)\cdot f^t,
\label{equ:ema}
\end{equation}
where $\mathcal{M}[i]^t$ and $\mathcal{M}[i]^{t-1}$ respectively refer to the memory bank representations in the $t$ and $t-1$ epochs. The memory bank stores moving averaged representations, which stabilize the pseudo label generation. To further enhance the pseudo label quality, we compute k-reciprocal re-ranked Jaccard distance~\cite{zhong2017re} between memory bank representations, which are then fed into a clustering algorithm DBSCAN~\cite{Ester1996ADA} to generate pseudo labels $\mathcal{Y}=\{y_1, y_2, ..., y_N \}$. 
During the training, the pseudo labels are renewed at the beginning of each epoch. We design a Rotation Contrast and a Mixup Contrast respectively for the two types of generated views.

\begin{table}
\caption{Performance comparison of rotating one mesh and switching two random meshes in the generation.}
\scalebox{1}{
\begin{tabular}{c|cc|cc}
\hline
\multirow{2}{*}{Method}  & \multicolumn{2}{c}{Duke$\to$Market} & \multicolumn{2}{|c}{Market$\to$Duke} \\ \cline{2-5}
\multicolumn{1}{c|}{} & \multicolumn{1}{c}{mAP} & \multicolumn{1}{c|}{Rank1} & \multicolumn{1}{c}{mAP} & \multicolumn{1}{c}{Rank1} \\ \hline
2D gray image switch~\cite{zheng2019joint}&60.1&78.8&59.5&76.2\\
Mesh switch&74.2&88.5&60.6&76.9\\
Mesh rotation&\textbf{74.4}&\textbf{89.7}&\textbf{61.3}&\textbf{78.0}\\
\hline
\end{tabular}}
\centering
\label{table:mesh rotation mesh switch}
\end{table}

\subsubsection{Rotation Contrast (for id-unrelated augmentation)}
As shown in Fig.~\ref{fig:general structure}.~(d), the original image $x$ and the generated image $x_{new}^{\prime}$ are encoded by the shared identity encoder into two identity feature vectors $E_{id}(x)\to f$ and $E_{id}(x_{new}^{\prime})\to f_{new}^{\prime}$. For a representation $f$ with a pseudo label $y_i$, we randomly sample a positive representation $f_{pos}$ of the same pseudo label $y_i$ and $K$ negative representations of pseudo labels different to $y_i$ from the memory bank. Three positive pairs can be formed, \ie, $(f, f_{pos})$, $(f, f_{new}^{\prime})$ and $(f_{pos}, f_{new}^{\prime})$. The $f_{new}^{\prime}$ and sampled $K$ negative representations from the memory bank are used to form $K$ negative pairs. We define three view-invariant losses to attract three positive pairs while repulsing $K$ negative pairs:
\begin{equation}
\begin{split}
  \mathcal{L}_{vi} = \mathop{\mathbb{E}}[\log{(1+\frac{\sum\nolimits_{i=1}^{K}\exp{(<f_{new}^{\prime}\cdot k_{i}>/\tau)}}{\exp{(<f\cdot f_{pos}>/\tau)}})}],
\label{view invariant loss}
\end{split}
\end{equation}
\begin{equation}
\begin{split}
  \mathcal{L}_{vi}^{\prime} = \mathop{\mathbb{E}}[\log{(1+\frac{\sum\nolimits_{i=1}^{K}\exp{(<f_{new}^{\prime}\cdot k_{i}>/\tau)}}{\exp{(<f_{new}^{\prime}\cdot f>/\tau)}})}],
\label{view invariant loss prime}
\end{split}
\end{equation}
\begin{equation}
\begin{split}
  \mathcal{L}_{vi}^{\prime\prime} = \mathop{\mathbb{E}}[\log{(1+\frac{\sum\nolimits_{i=1}^{K}\exp{(<f_{new}^{\prime}\cdot k_{i}>/\tau)}}{\exp{(<f_{new}^{\prime}\cdot f_{pos}>/\tau)}})}],
\label{view invariant loss prime prime}
\end{split}
\end{equation}
where $<\cdot >$ denotes the cosine similarity between two feature vectors. $\tau$ is a temperature hyper-parameter to sharpen the cosine similarity. $k_{i}$ denotes negative representations sampled from the memory bank. Presented three loss functions enable the contrastive module to maximize the similarity between original view $f$, generated view $f_{new}^{\prime}$ and positive memory view $f_{pos}$. At the same time, the similarity between generated view $f_{new}^{\prime}$ and $K$ negative memory views is minimized, which encourages the generative module to refine the generated view $f_{new}^{\prime}$ that should be different from a large number of negative samples.

\subsubsection{Mixup Contrast (for id-related augmentation)}
\begin{figure}
\centering
\includegraphics[width=1\linewidth]{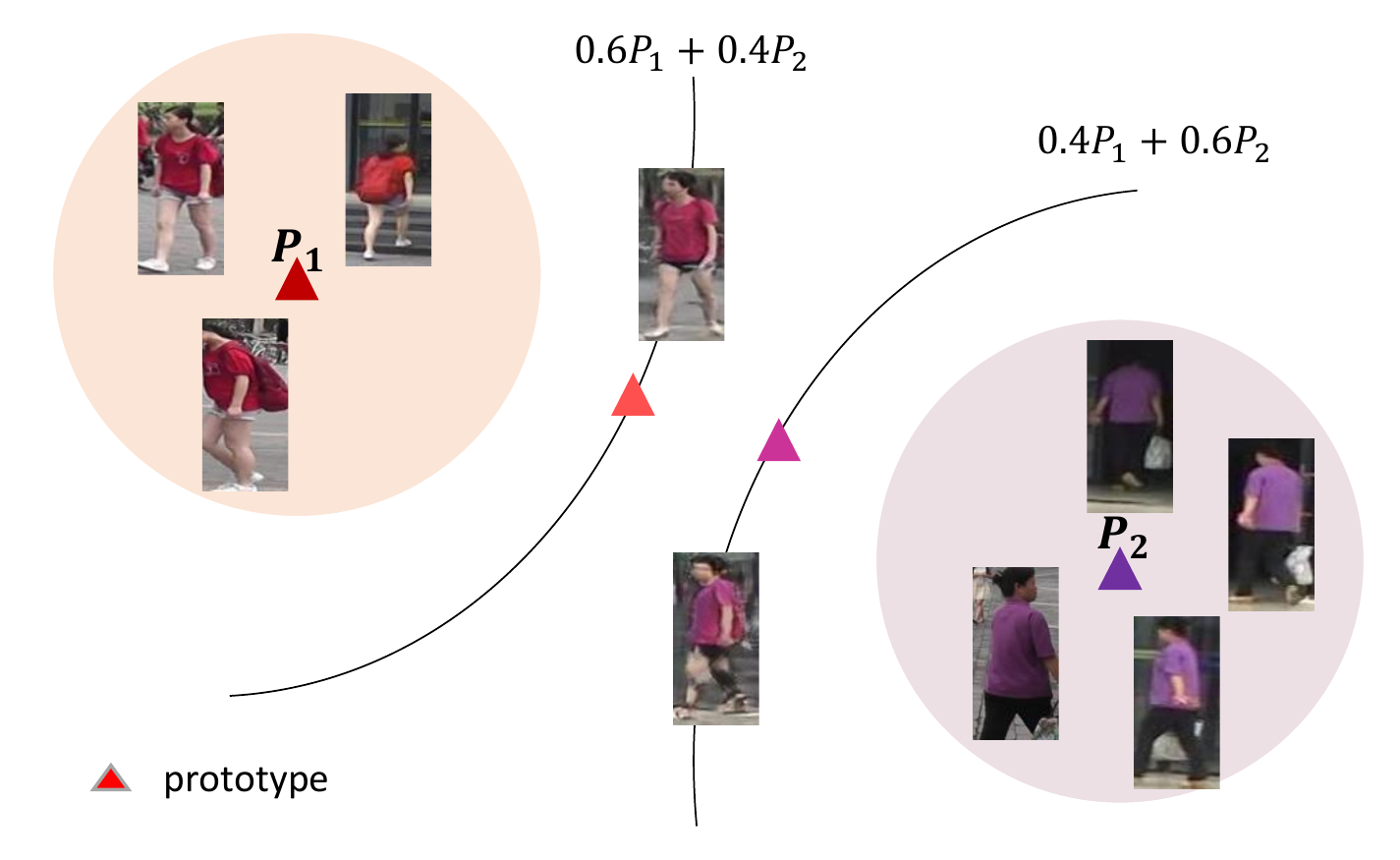}
\caption{Mixup Contrast targets at learning a smoother decision boundary between two persons $P_1$ and $P_2$ by contrasting in-between samples with in-between prototypes.}
\label{fig_picture3_contrast_mixup}
\end{figure}

The mixed image $x_{mix}^{\prime}$ is encoded by the shared identity encoder into a mixed identity feature vector $E_{id}(x_{mix}^{\prime}) \to f_{mix}^{\prime}$, see Fig.~\ref{fig:general structure}.~(e). 
Towards learning a smoother decision boundary between two clusters, as illustrated in Fig.~\ref{fig_picture3_contrast_mixup}, we design a Mixup Contrast for $f_{mix}^{\prime}$. As certain instances in a cluster are close to the decision boundary between two clusters, whereas the others are far away, we define an averaged prototype for a cluster:
\begin{equation}
   p^{}_{a} = \frac{1}{N_{a}}\sum_{ \mathcal{M}[i] \in y_{a}} \mathcal{M}[i],
\label{equa:negative prototype}
\end{equation}
where $N_a$ is the number of instances belonging to the cluster $a$.

Given a random image representation $f$, we use a softmax cross-entropy loss $\mathcal{L}_{proto}$ to make $f$ converge to the cluster prototype, which encourages the compactness of a cluster:
\begin{equation}
\begin{split}
  \mathcal{L}_{proto} = \mathop{\mathbb{E}}[\log{(1+\frac{\sum\nolimits_{i=1}^{|\mathcal{Y}|-1}\exp{(f_{}^{}\cdot p_{i})}}{\exp{(f_{}^{}\cdot p_{+})}})}],
\label{prototype contrast}
\end{split}
\end{equation}
where $p_{+}$ is the corresponding prototype of $f$ and $p_{i}$ denotes other cluster prototypes. $|\mathcal{Y}|$ is the number of clusters. Given that certain clusters may contain more instances that are close to decision boundaries with other clusters, compact clusters provide stable mixed prototypes. 

Based on the pseudo labels, we define a mixed prototype vector between two clusters $i$ and $j$:
\begin{equation}
   p^{}_{mix} = \lambda^* \cdot p^{}_{i} +(1-\lambda^*)\cdot p^{}_{j},
\label{equa:positive prototype}
\end{equation}
where $\lambda^*$ is the same mixing coefficient as in Eq.~(\ref{mixup_feat}). 

For the mixed representation $f_{mix}^{\prime}$, we use another softmax cross-entropy loss to maximize its similarity with the mixed prototype $p_{mix}$ and minimize its similarity with $|\mathcal{Y}|-2$ negative prototypes that do not belong to the two clusters $i$ and $j$:
\begin{equation}
\begin{split}
  \mathcal{L}_{mix} = \mathop{\mathbb{E}}[\log{(1+\frac{\sum\nolimits_{i=1}^{|\mathcal{Y}|-2}\exp{(f_{mix}^{\prime}\cdot p_{i})}}{\exp{(f_{mix}^{\prime}\cdot p_{mix})}})}].
\label{mixup contrast}
\end{split}
\end{equation}
As opposed to cosine similarity in Eq.~(\ref{view invariant loss}), (\ref{view invariant loss prime}) and (\ref{view invariant loss prime prime}), we do not compute normalized similarity, as the average operation for computing prototype vectors performs as normalization.

\subsubsection{Overall contrastive loss}
The overall contrastive loss combines the above losses (\ref{view invariant loss}), (\ref{view invariant loss prime}), (\ref{view invariant loss prime prime}), (\ref{prototype contrast}) and (\ref{mixup contrast}):
\begin{equation}
   \mathcal{L}_{contrast} = \lambda_{vi}(\mathcal{L}_{vi} +  \mathcal{L}_{vi}^{\prime} +  \mathcal{L}_{vi}^{\prime\prime}) +\lambda_{mix}(\mathcal{L}_{proto} + \mathcal{L}_{mix}).
\label{overall contrast loss}
\end{equation}

\subsection{Joint Training}
Our proposed framework incorporates a generative module and a contrastive module. The generative module disentangles a person image representation into identity and structure features, which allows for learning purified identity features for person ReID. The contrastive module learns invariance via contrasting augmented images. If we replace the GAN-based augmentation with traditional data augmentation techniques, both modules can be trained separately. However, a separate training leads to sub-optimal performance for both of them. To address this issue, we couple the two modules with a shared identity encoder in a joint training framework. In the setting of joint training, both modules work collaboratively to achieve one objective: enhancing the discriminality of identity representations. Inside GCL+, the generative module provides both, id-unrelated and id-related augmentations for the contrastive module. On the other hand, the contrastive module maximizes the similarity between positive views, while repulsing negative views, which, in turn, refines the identity representations for a better generation quality. Both modules mutually promote each other's performance in the joint training, leading to an optimal ReID performance. 
In our proposed framework, a forward propagation is firstly conducted on the generative module and subsequently on the contrastive module.
A backward propagation is then conducted with an overall loss that combines Eq.~(\ref{overall_gan_loss}) and Eq.~(\ref{overall contrast loss}):
\begin{equation}
   \mathcal{L}_{overall} = \mathcal{L}_{gan} + \mathcal{L}_{contrast}.
\label{overall loss}
\end{equation}

\section{Experiment}
\subsection{Datasets and Evaluation Protocols}
We evaluate our proposed method GCL+ on five mainstream person ReID benchmarks, including three image-based datasets: Market-1501~\cite{Zheng2015ScalablePR}, DukeMTMC-reID~\cite{ristani2016MTMC}, MSMT17~\cite{wei2018person} and two video-based datasets: MARS~\cite{zheng2016mars} and DukeMTMC-VideoReID~\cite{Wu2018ExploitTU}. \textbf{Market-1501} dataset is collected in front of a supermarket in Tsinghua University from 6 cameras. It is composed of 12,936 images of 751 identities for training and 19,732 images of 750 identities for testing. \textbf{DukeMTMC-reID} is collected from 8 cameras installed in the campus of Duke University. It contains 16,522 images of 702 persons for training, 2,228 query images and 17,661 gallery images of 702 persons for testing. \textbf{MSMT17} is a large-scale Re-ID dataset, which includes 32,621 training images of 1,041 identities and 93,820 testing images of 3,060 identities collected from 15 cameras deployed in both indoor and outdoor scenes. \textbf{MARS} is a large-scale video-based person ReID dataset. The dataset contains 17,503 tracklets of 1,261 identities collected from 6 cameras, where 625 identities are used for training and the other 636 identities are used for testing. \textbf{DukeMTMC-VideoReID} is a video-based person ReID dataset derived from DukeMTMC~\cite{ristani2016MTMC} dataset. DukeMTMC-VideoReID contains 2,196 training tracklets of
702 identities and 2,636 testing tracklets of other 702 identities.

As our method includes a GAN and a contrastive module, we report results for both unsupervised person ReID and generation quality evaluations. For unsupervised person ReID evaluation, we provide results under both, unsupervised domain adaptation and fully unsupervised settings. We report both, Cumulative Matching Characteristics (CMC) at Rank1, Rank5, Rank10 accuracies, as well as mean Average Precision (mAP) on the testing set. For the generation quality evaluation, we conduct a qualitative comparison between our method and state-of-the-art methods on generated images.

\subsection{Implementation details}
We introduce implementation details pertained to network design and general training configurations, as well as three-step optimization. 

\textbf{Network design.}
Our network design related to the identity encoder $E_{id}$, the structure encoder $E_{str}$, the decoder $G$ and the discriminator $D$ has been mainly inspired by \cite{zheng2019joint,Zou2020JointDA}. In the following descriptions, we denote the size of feature maps in channel$\times$height$\times$width.  \textbf{1)} $E_{id}$ is an ImageNet \cite{Russakovsky2015ImageNetLS} pre-trained ResNet50 \cite{he2016deep} with slight modifications. The original fully connected layer is replaced by a batch normalization layer and a fully connected embedding layer, which outputs identity representations $f$ in 512$\times$1$\times$1 for the contrastive module. In parallel, we add a part average pooling that outputs identity features $f_{id}$ in 2048$\times$4$\times$1 for the generative module. \textbf{2)} $E_{str}$ is composed of four convolutional and four residual layers, which output structure features $f_{str}$ in 128$\times$64$\times$32. \textbf{3)} $G$ contains four residual and four convolutional layers. Every residual layer contains two adaptive instance normalization layers \cite{huang2017arbitrary} that transform $f_{id}$ into scale and bias parameters. \textbf{4)} $D$ is a multi-scale PatchGAN \cite{isola2017image} discriminator at 64$\times$32, 128$\times$64 and 256$\times$128. 

\textbf{General training configurations.} 
Our framework is implemented under Pytorch~\cite{Paszke2019PyTorchAI} and trained with one Nvidia V100 GPU. The inputs are resized to 256$\times$128. We empirically set a large weight $\lambda_{recon}=5$ for reconstruction in Eq.~(\ref{overall_gan_loss}). With a batch size of 16, we use SGD to train $E_{id}$ and Adam optimizer to train $E_{str}$, $G$ and $D$. Learning rate in Adam is set to $1\times 10^{-4}$ and $3.5\times10^{-4}$ in SGD and are multiplied by $0.1$ after $10$ epochs. DBSCAN maximal neighborhood distance is set to $0.5$ and minimal sample number is set to 4. The number of negatives $K$ is 8192. For testing, $E_{id}$ outputs representations $f$ of dimension 512. For video-based person ReID, due to our GPU memory constraint, we randomly sample 2 frames per tracklet on MARS and 8 frames per tracklet on DukeMTMC-VideoReID for training. For testing, all the frames from each tracklet are used to calculate a unified tracklet representation for similarity ranking. Other settings are kept the same as image-based peron ReID settings. 

\textbf{Three-stage optimization.} 
To reduce the noise from imperfect generated images at early epochs, we train the four modules $E_{id}$, $E_{str}$, $G$ and $D$ in a three-stage optimization. \textbf{Stage 1} $E_{id}$ warm-up: we use a state-of-the-art unsupervised ReID method to warm up $E_{id}$, \eg, ACT~\cite{yang2020asymmetric}, MMCL~\cite{Wang_2020_CVPR} and JVTC~\cite{li2020joint}. \textbf{Stage 2} $E_{str}$, $G$ and $D$ warm-up: we freeze $E_{id}$ and warm up $E_{str}$, $G$, and $D$ only with the overall GAN loss in Eq.~(\ref{overall_gan_loss}) for 40 epochs. \textbf{Stage 3} joint training: we bring in the memory bank and the pseudo labels to jointly train the whole framework with the overall loss in Eq.~(\ref{overall loss}) for another 20 epochs. 

\subsection{Unsupervised ReID Evaluation}
To validate the effectiveness of each component, we conduct parameter analysis and ablation experiments with a JVTC~\cite{li2020joint} baseline. As JVTC+ is the enhanced version of JVTC with a camera temporal distribution post-processing, the performance boost from the post-processing is almost fixed. Thus, the ablation experiments show similar variance with JVTC and JVTC+ baselines. We further compare our method with state-of-the-art unsupervised person ReID with three different baselines to show the generalizability of our method.

\begin{figure}
\centering
   \includegraphics[width=0.95\linewidth]{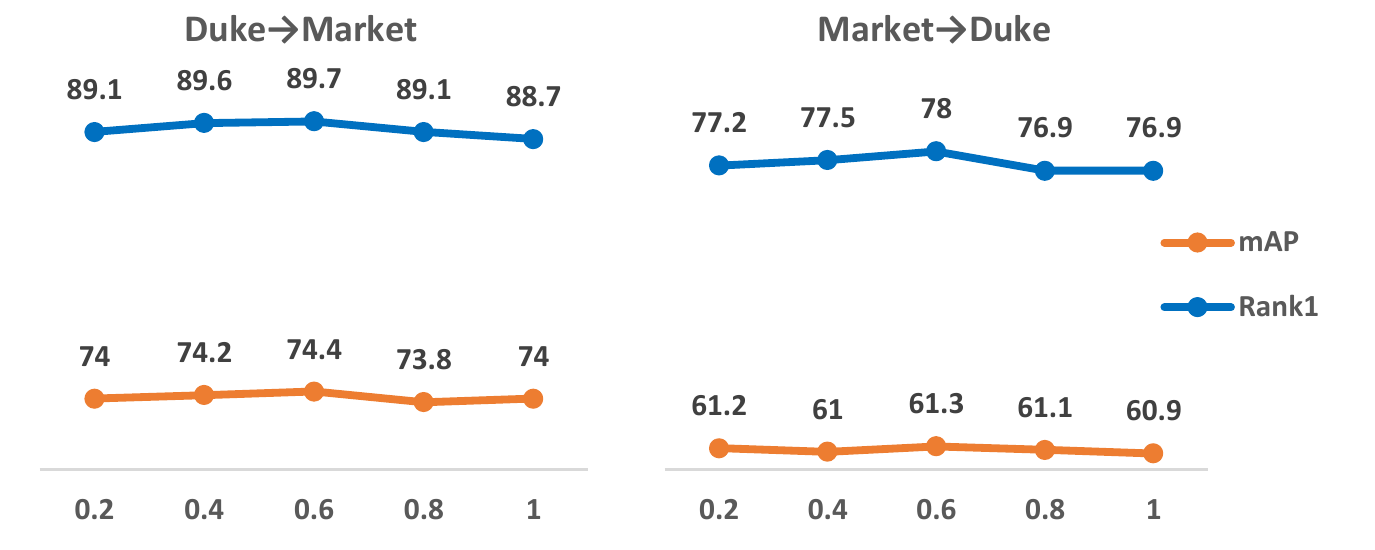}
   \caption{Hyper-parameter analysis on $\alpha$ for mixup coefficient on Duke$\to$Market and Market$\to$Duke tasks. }
\label{fig:alpha}
\end{figure}

\begin{figure}
\centering
   \includegraphics[width=0.95\linewidth]{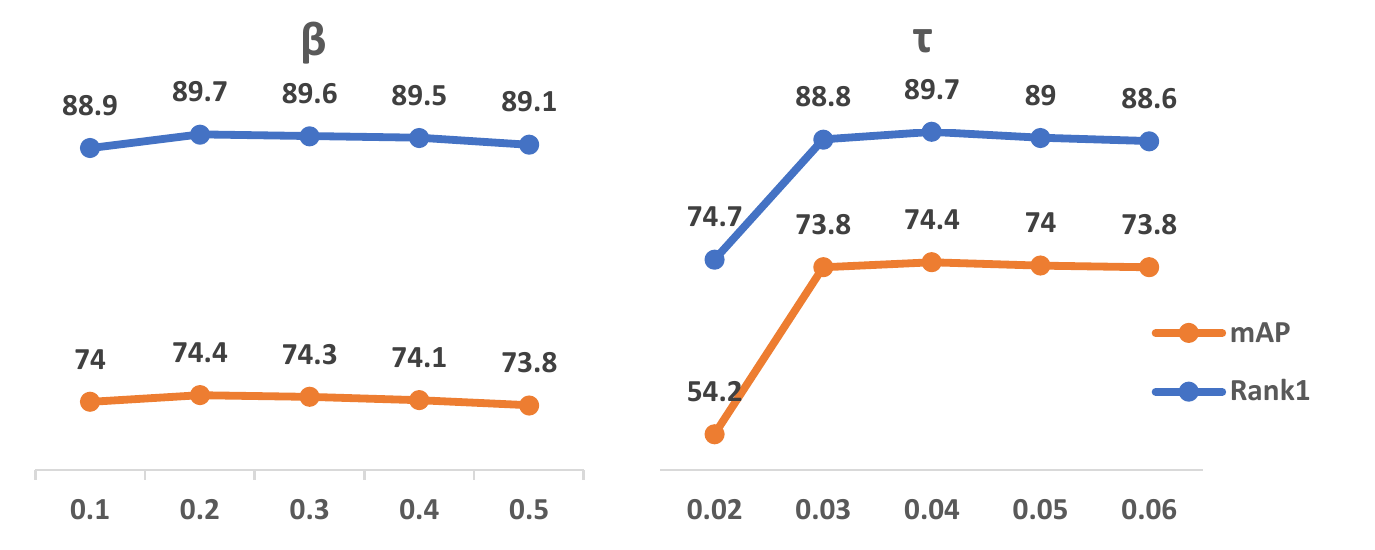}
   \caption{Hyper-parameter analysis on $\beta$ for memory momentum and $\tau$ for contrastive temperature on Duke$\to$Market task. }
\label{fig:beta and tau}
\end{figure}

\begin{figure}
\centering
\includegraphics[width=1\linewidth]{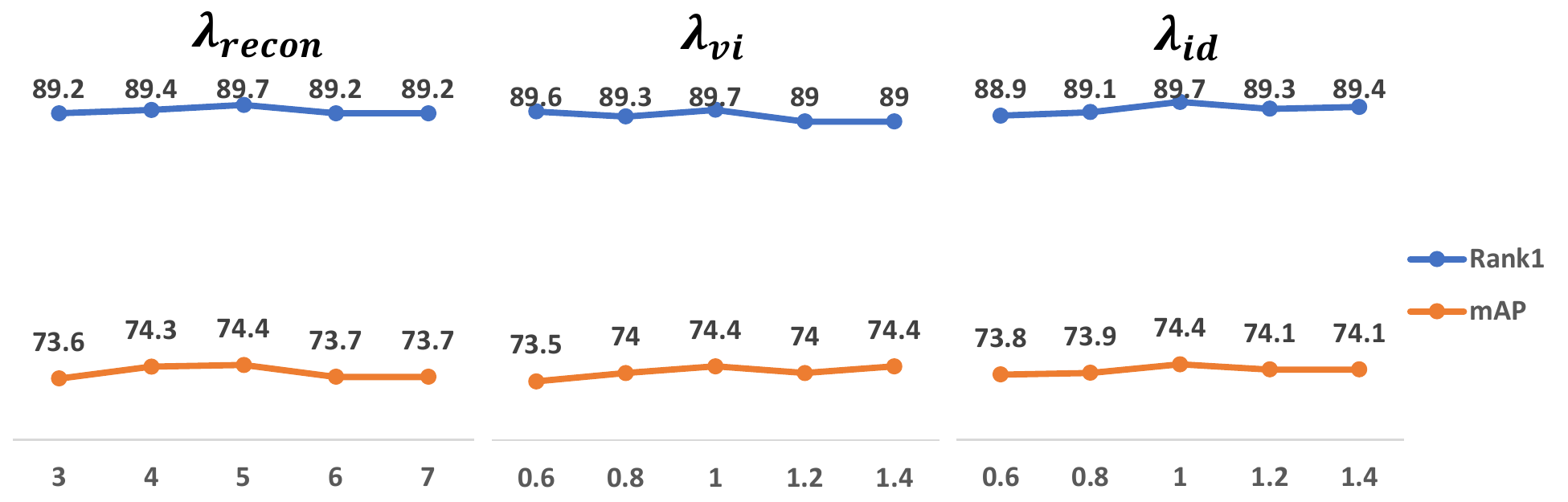}
\caption{Hyper-parameter analysis on balancing coefficients  $\lambda_{recon}$ for reconstruction weight, $\lambda_{vi}$ for rotation contrast weight and $\lambda_{mix}$ for mixup contrast weight on Duke→Market task.}
\label{fig_balancing_coefficients}
\end{figure}

\begin{table}
\caption{Performance under different clustering neighborhood distance threshold. `N' is the approximate number of pseudo-identities.}
\scalebox{0.9}{
\begin{tabular}{c|ccc|ccc}
\hline
\multirow{2}{*}{Threshold}  & \multicolumn{3}{c}{Duke$\to$Market} & \multicolumn{3}{|c}{Market$\to$Duke} \\ \cline{2-7}
\multicolumn{1}{c|}{} & \multicolumn{1}{c}{N} & \multicolumn{1}{c}{mAP} & \multicolumn{1}{c|}{Rank1} & \multicolumn{1}{c}{N} & \multicolumn{1}{c}{mAP} & \multicolumn{1}{c}{Rank1} \\ \hline
0.4&$\sim$642&\textbf{74.5}&89.4&$\sim$840&60.9&77.1\\
0.45&$\sim$605&74.4&89.4&$\sim$810&61.2&77.4\\
0.5&$\sim$584&74.4&\textbf{89.7}&$\sim$786&\textbf{61.3}&\textbf{78.0}\\
0.55&$\sim$540&73.6&88.4&$\sim$744&61.1&76.8\\
0.6&$\sim$500&72.4&87.6&$\sim$697&60.7&77.7\\
\hline
\end{tabular}}
\centering
\label{table:threshold}
\end{table}

\begin{table*}
\centering
\caption{Ablation study under fully unsupervised and UDA settings on traditional (w/o GAN) and generative (w/ GAN) data augmentation for the contrastive module. `Multi' refers to multiple commonly used data augmentation techniques for person ReID, including random flipping, padding, cropping and erasing. `Rotation' refers to our proposed mesh-guided rotation. `Mixup' is conducted on image level, while `F-Mixup' is conducted on feature level. }
\scalebox{0.9}{
\begin{tabular}{c|ccccc|cccc|cccc}
\hline
\multirow{2}{*}{Fully unsupervised}&\multicolumn{2}{c}{ID-unrelated} &\multicolumn{3}{|c}{ID-related} & \multicolumn{4}{|c}{Market} & \multicolumn{4}{|c}{Duke} \\ \cline{2-14}
&\multicolumn{1}{c}{Multi} &\multicolumn{1}{c}{Rotation} &\multicolumn{1}{|c}{Mixup} &\multicolumn{1}{c}{F-Mixup} &\multicolumn{1}{c}{D-Mixup} & \multicolumn{1}{|c}{mAP} & \multicolumn{1}{c}{R1} & \multicolumn{1}{c}{R5}& \multicolumn{1}{c}{R10}& \multicolumn{1}{|c}{mAP} & \multicolumn{1}{c}{R1}& \multicolumn{1}{c}{R5}& \multicolumn{1}{c}{R10} \\ \hline
\multirow{3}{*}{w/o GAN}&\multicolumn{5}{c|}{Baseline}&47.2&75.4&86.7&90.5&43.9&66.8&77.6&81.0\\
&\checkmark&&&&&58.2&81.1&91.0&93.5&50.8&70.8&80.9&83.8\\
&\checkmark&&\checkmark&&&60.0&82.5&91.6&94.0&51.0&71.1&80.8&84.1\\
\hline
\multirow{4}{*}{w/ GAN}&&\checkmark&&&&63.8&83.4&91.8&94.3&53.1&72.8&81.2&83.7\\
&&\checkmark&\checkmark&&&65.9&84.8&92.5&94.3&54.3&73.6&82.5&84.9\\
&&\checkmark&&\checkmark&&66.1&84.3&92.4&94.6&54.2&73.7&82.4&85.5\\
&&\checkmark&&&\checkmark&\textbf{66.3}&\textbf{85.3}&\textbf{92.9}&\textbf{94.6}&\textbf{54.6}&\textbf{74.2}&\textbf{82.8}&\textbf{85.6}\\
\hline
\hline
\multirow{2}{*}{UDA}&\multicolumn{2}{c}{ID-unrelated} &\multicolumn{3}{|c}{ID-related} & \multicolumn{4}{|c}{Duke$\to$Market} & \multicolumn{4}{|c}{Market$\to$Duke} \\ \cline{2-14}
&\multicolumn{1}{c}{Multi} &\multicolumn{1}{c}{Rotation} &\multicolumn{1}{|c}{Mixup} &\multicolumn{1}{c}{F-Mixup} &\multicolumn{1}{c}{D-Mixup} & \multicolumn{1}{|c}{mAP} & \multicolumn{1}{c}{R1} & \multicolumn{1}{c}{R5}& \multicolumn{1}{c}{R10}& \multicolumn{1}{|c}{mAP} & \multicolumn{1}{c}{R1}& \multicolumn{1}{c}{R5}& \multicolumn{1}{c}{R10} \\ \hline
\multirow{3}{*}{w/o GAN}&\multicolumn{5}{c|}{Baseline}&65.0&85.7&93.4&95.9&56.5&73.9&84.4&87.8\\
&\checkmark&&&&&70.4&86.9&94.3&95.8&57.0&74.2&84.2&87.2\\
&\checkmark&&\checkmark&&&70.7&87.8&94.1&96.3&57.7&74.5&85.0&88.0\\
\hline
\multirow{4}{*}{w/ GAN}&&\checkmark&&&&72.5&88.7&94.8&96.3&59.9&75.9&86.2&88.5\\
&&\checkmark&\checkmark&&&73.0&88.9&94.8&96.4&60.4&76.5&85.9&88.3\\
&&\checkmark&&\checkmark&&72.7&88.8&95.1&96.3&60.2&76.7&86.1&88.1\\
&&\checkmark&&&\checkmark&\textbf{74.4}&\textbf{89.7}&\textbf{95.5}&\textbf{96.7}&\textbf{61.3}&\textbf{78.0}&\textbf{86.8}&\textbf{89.1}\\
\hline
\end{tabular}}
\label{table:fully unsupervised ablation}
\end{table*}

\begin{table}
\centering
\caption{Ablation study on three view-invariant losses in Rotation Contrast and two prototype losses in Mixup Contrast.}
\scalebox{0.91}{
\begin{tabular}{ccccc|cc|cc}
\hline
\multirow{2}{*}{$\mathcal{L}_{vi}$}  & \multirow{2}{*}{$\mathcal{L}_{vi}^{\prime}$} & \multirow{2}{*}{$\mathcal{L}_{vi}^{\prime\prime}$} & \multirow{2}{*}{$\mathcal{L}_{proto}$} & \multirow{2}{*}{$\mathcal{L}_{mix}$} & \multicolumn{2}{c}{Duke$\to$Market} & \multicolumn{2}{|c}{Market$\to$Duke} \\ \cline{6-9}
\multicolumn{1}{c}{} &\multicolumn{1}{c}{} &\multicolumn{1}{c}{}&\multicolumn{1}{c}{}&\multicolumn{1}{c}{} & \multicolumn{1}{|c}{mAP} & \multicolumn{1}{c|}{R1} & \multicolumn{1}{c}{mAP} & \multicolumn{1}{c}{R1} \\ \hline
\checkmark&&&&&61.6&82.4&51.7&70.6\\
\checkmark&\checkmark&&&&69.1&85.6&58.3&74.8\\
\checkmark&\checkmark&\checkmark &&&72.5&88.7&59.9&75.9\\
\checkmark&\checkmark&\checkmark &\checkmark&&72.8&88.8&60.6&76.9\\
\checkmark&\checkmark&\checkmark &\checkmark&\checkmark&\textbf{74.4}&\textbf{89.7}&\textbf{61.3}&\textbf{78.0}\\
\hline
\end{tabular}}
\label{table:vi vi2 vi3}
\end{table}

\subsubsection{Parameter analysis}
Hyper-parameters, such as mixing coefficient $\alpha$, memory momentum $\beta$ and view-invariant contrastive loss temperature $\tau$, play important roles inside our proposed GCL+ framework for better unsupervised person ReID performance. We vary their values to analyze the sensitivity of each hyper-parameter inside our proposed framework GCL+. 

For Beta distribution, a larger $\alpha$ results in a higher possibility that $\lambda$ gets closer to 0.5. ReID performance on both Duke$\to$Market and Market$\to$Duke tasks with reference to $\alpha$ is reported in Fig.~\ref{fig:alpha}. On both tasks, the optimal performance is achieved, in case of $\alpha$ is around 0.6. As a consequence, $\alpha$ is set to 0.6 in our framework.

The value of $\beta$ controls the memory updating speed. The value of $\tau$ amplifies the cosine similarity between contrastive views. An overlarge or undersized value, generally speaking, introduces more noise for contrastive learning. We report the performance variation with reference to $\beta$ and $\tau$ on Duke$\to$Market task in Fig.~\ref{fig:beta and tau}. We find that the performance is more sensitive to the similarity temperature $\tau$. Based on the results, we set $\beta$ to 0.2 and $\tau$ to 0.04.

The number of possible pseudo-identities $N$ is related to clustering hyper-parameters, such as maximal neighborhood distance threshold and minimal cluster sample number. The distance threshold of DBSCAN is the maximal distance between two samples for one to be considered as in the neighborhood of the other. A larger distance threshold enlarges the radius of a cluster, making more samples be considered into a same cluster ($N$ becomes smaller). As shown in Table~\ref{table:threshold}, the threshold value only slightly affects ReID performance.

As our framework jointly optimize the generative and contrastive modules, we set weight coefficients to balance different loss functions in the two modules. We vary the balancing coefficients $\lambda_{recon}$, $\lambda_{vi}$ and $\lambda_{mix}$ in Equation~(\mbox{\ref{overall_gan_loss}}) and (\mbox{\ref{overall contrast loss}}). The corresponding results are reported in Fig.~\mbox{\ref{fig_balancing_coefficients}}. Overall, the different values in the tested range only slightly influence the final results. Based on the results, we set $\lambda_{recon}=5$, $\lambda_{vi}=1$ and $\lambda_{mix}=1$.

\begin{table*}
\caption{Comparison of fully unsupervised ReID methods (\%) on Market1501, DukeMTMC-reID and MSMT17 datasets. We test our proposed method on several baselines, see names in parentheses.} 
\label{table:fully unsupervised comparison}
\centering
\scalebox{0.94}{
\begin{tabular}{l|c|cccc|cccc|cccc}
\hline
\multirow{2}{*}{Method} & \multirow{2}{*}{Reference} & \multicolumn{4}{c}{Market1501} & \multicolumn{4}{|c}{DukeMTMC-reID} & \multicolumn{4}{|c}{MSMT17} \\ \cline{3-14}
\multicolumn{1}{c|}{} &\multicolumn{1}{c|}{}& \multicolumn{1}{c}{mAP} & \multicolumn{1}{c}{R1} & \multicolumn{1}{c}{R5} & \multicolumn{1}{c|}{R10} & \multicolumn{1}{c}{mAP} & \multicolumn{1}{c}{R1}  & \multicolumn{1}{c}{R5} & \multicolumn{1}{c|}{R10}& \multicolumn{1}{c}{mAP} & \multicolumn{1}{c}{R1} & \multicolumn{1}{c}{R5}& \multicolumn{1}{c}{R10}\\ 
\hline
BUC \cite{Lin2019ABC} &AAAI'19 &29.6&61.9&73.5&78.2&22.1&40.4&52.5&58.2&-&-&-&-\\ 
SoftSim \cite{Lin2020UnsupervisedPR} &CVPR'20 &37.8&71.7&83.8&87.4&28.6&52.5&63.5&68.9&-&-&-&-\\
TSSL \cite{wu2020tracklet} &AAAI'20&43.3&71.2&-&-&38.5&62.2&-&-&-&-&-&-\\
MMCL \cite{Wang_2020_CVPR} &CVPR'20 &45.5&80.3&89.4&92.3&40.2&65.2&75.9&80.0&11.2&35.4&44.8&49.8\\
JVTC \cite{li2020joint}&ECCV'20&41.8&72.9&84.2&88.7&42.2&67.6&78.0&81.6&15.1&39.0&50.9&56.8\\
JVTC+ \cite{li2020joint}&ECCV'20&47.5&79.5&89.2&91.9&50.7&74.6&82.9&85.3&17.3&43.1&53.8&59.4\\
MetaCam \cite{yang2021joint}&CVPR'21&61.7&83.9&92.3&-&53.8&73.8&84.2&-&15.5&35.2&48.3&-\\
\hline
GCL(MMCL) \cite{Chen_2021_joint}&CVPR'21&54.9&83.7&91.6&94.0&49.3&69.7&79.7&82.8&-&-&-&-\\
GCL(JVTC) \cite{Chen_2021_joint}&CVPR'21&63.4&83.7&91.6&94.3&53.3&72.4&82.0&84.9&18.0&41.6&53.2&58.4\\
GCL(JVTC+) \cite{Chen_2021_joint}&CVPR'21&66.8&87.3&93.5&95.5&62.8&82.9&87.1&88.5&21.3&45.7&58.6&64.5\\
\hline
GCL+(MMCL) &This paper&56.0&84.0&91.4&93.7&49.5&70.2&80.2&83.3&-&-&-&-\\
GCL+(JVTC) &This paper&66.3&85.3&92.9&94.6&54.6&74.2&82.8&85.6&19.2&44.7&56.4&61.4\\
GCL+(JVTC+) &This paper&\textbf{69.3}&\textbf{89.0}&\textbf{94.6}&\textbf{96.0}&\textbf{63.5}&\textbf{83.1}&\textbf{87.4}&\textbf{88.8}&\textbf{22.0}&\textbf{47.9}&\textbf{61.3}&\textbf{67.1}\\
\hline
\end{tabular}}
\end{table*}

\begin{table*}
\caption{Comparison of unsupervised domain adaptive ReID methods (\%) between Market1501, DukeMTMC-reID and MSMT17 datasets. We test our proposed method on several baselines, see names in parentheses.} 
\label{table:UDA comparison}
\centering
\scalebox{0.9}{
\begin{tabular}{l|c|cccc|cccc|cccc|cccc}
\hline
\multirow{2}{*}{Method} & \multirow{2}{*}{Reference} & \multicolumn{4}{c}{Duke$\to$Market} & \multicolumn{4}{|c}{Market$\to$Duke} & \multicolumn{4}{|c}{Market$\to$MSMT17} & \multicolumn{4}{|c}{Duke$\to$MSMT17} \\ \cline{3-18}
\multicolumn{1}{c|}{} &\multicolumn{1}{c|}{}& \multicolumn{1}{c}{mAP} & \multicolumn{1}{c}{R1} & \multicolumn{1}{c}{R5} & \multicolumn{1}{c|}{R10} & \multicolumn{1}{c}{mAP} & \multicolumn{1}{c}{R1}  & \multicolumn{1}{c}{R5} & \multicolumn{1}{c|}{R10}& \multicolumn{1}{c}{mAP} & \multicolumn{1}{c}{R1} & \multicolumn{1}{c}{R5}& \multicolumn{1}{c|}{R10}& \multicolumn{1}{c}{mAP} & \multicolumn{1}{c}{R1} & \multicolumn{1}{c}{R5}& \multicolumn{1}{c}{R10}\\ 
\hline
ECN \cite{zhong2019invariance}&CVPR'19&43.0&75.1&87.6&91.6&40.4&63.3&75.8&80.4&8.5&25.3&36.3&42.1&10.2&30.2&41.5&46.8\\
PDA \cite{li2019cross}&ICCV'19&47.6&75.2&86.3&90.2&45.1&63.2&77.0&82.5&-&-&-&-&-&-&-&-\\
CR-GAN \cite{chen2019instance}&ICCV'19&54.0&77.7&89.7&92.7&48.6&68.9&80.2&84.7&-&-&-&-&-&-&-&-\\
SSG \cite{fu2019self}&ICCV'19&58.3&80.0&90.0&92.4&53.4&73.0&80.6&83.2&13.2&31.6&49.6&-&13.3&32.2&51.2&-\\
MMCL \cite{Wang_2020_CVPR} &CVPR'20 &60.4&84.4&92.8&95.0&51.4&72.4&82.9&85.0&15.1&40.8&51.8&56.7&16.2&43.6&54.3&58.9\\
ACT \cite{yang2020asymmetric}&AAAI'20&60.6&80.5&-&-&54.5&72.4&-&-&-&-&-&-&-&-&-&-\\
DG-Net++ \cite{Zou2020JointDA}&ECCV'20&61.7&82.1&90.2&92.7&63.8&78.9&87.8&90.4&22.1&48.4&60.9&66.1&22.1&48.8&60.9&65.9\\
JVTC \cite{li2020joint}&ECCV'20&61.1&83.8&93.0&95.2&56.2&75.0&85.1&88.2&19.0&42.1&53.4&58.9&20.3&45.4&58.4&64.3\\
ECN+ \cite{zhong2020learning}&TPAMI'20&63.8&84.1&92.8&95.4&54.4&74.0&83.7&87.4&15.2&40.4&53.1&58.7&16.0&42.5&55.9&61.5\\
JVTC+ \cite{li2020joint}&ECCV'20&67.2&86.8&95.2&97.1&66.5&80.4&89.9&92.2&25.1&48.6&65.3&68.2&27.5&52.9&70.5&75.9\\
MMT \cite{ge2020mutual}&ICLR'20&71.2&87.7&94.9&96.9&65.1&78.0&88.8&92.5&22.9&49.2&63.1&68.8&23.3&50.1&63.9&69.8\\
CAIL \cite{luo2020generalizing}&ECCV'20&71.5&88.1&94.4&96.2&65.2&79.5&88.3&91.4&20.4&43.7&56.1&61.9&24.3&51.7&64.0&68.9\\
MetaCam \cite{yang2021joint}&CVPR'21&76.5&90.1&-&-&65.0&79.5&-&-&-&-&-&-&-&-&-&-\\
\hline
GCL(ACT) \cite{Chen_2021_joint}&CVPR'21&66.7&83.9&91.4&93.4&55.4&71.9&81.6&84.6&-&-&-&-&-&-&-&-\\
GCL(JVTC) \cite{Chen_2021_joint}&CVPR'21&73.4&89.1&95.0&96.6&60.4&77.2&86.2&88.4&21.5&45.0&57.1&66.5&24.9&50.8&63.4&68.9\\
GCL(JVTC+) \cite{Chen_2021_joint}&CVPR'21&75.4&90.5&96.2&97.1&67.6&81.9&88.9&90.6&27.0&51.1&63.9&69.9&29.7&54.4&68.2&74.2\\
\hline
GCL+(ACT) &This paper&67.5&84.3&92.6&94.2&56.8&73.5&82.8&85.1&-&-&-&-&-&-&-&-\\
GCL+(JVTC) &This paper&74.4&89.7&95.5&96.7&61.3&78.0&86.8&89.1&23.0&48.3&60.6&65.8&25.5&52.7&65.2&70.2\\
GCL+(JVTC+) &This paper&\textbf{76.5}&\textbf{91.6}&\textbf{96.3}&\textbf{97.6}&\textbf{68.3}&\textbf{82.6}&\textbf{89.4}&\textbf{91.2}&\textbf{27.8}&\textbf{53.8}&\textbf{66.9}&\textbf{72.5}&\textbf{31.5}&\textbf{57.9}&\textbf{70.3}&\textbf{76.1}\\
\hline
\end{tabular}}
\end{table*}

\subsubsection{Ablation study}
Contrastive learning methods strongly rely on data augmentation to create different augmented views for contrasting. Our proposed GCL+ outperforms traditional contrastive learning methods by replacing traditional data augmentation techniques with GAN-based augmentation techniques. To validate the effectiveness of our proposed GAN-based augmentation techniques and contrastive losses, we conduct ablation experiments on both Market-1501 and DukeMTMC-reID datasets. 

\textbf{Data augmentation.} Data augmentation techniques can be caterogized into id-unrelated and id-related augmentation. Id-unrelated augmentation creates intra-image visual distortions. In contrast, id-related augmentation creates inter-image visual distortions, which affects image identities. We compare results of traditional and generative data augmentation under fully unsupervised setting and domain adaptation setting in Table~\ref{table:fully unsupervised ablation}. For traditional data augmentation, we use multiple popular person ReID data augmentation techniques, including random flipping, padding, cropping and erasing~\cite{Zhong2020RandomED}, as id-unrelated augmentation and Mixup~\cite{zhang2018mixup} as id-related augmentation. Even with these traditional data augmentation, our contrastive module significantly outperforms the baseline. When we replace traditional data augmentation with generative data augmentation, the unsupervised person ReID performance can be further improved. Our proposed mesh-guided rotation (Rotation) works better than the multiple commonly used data augmentation techniques (Multi) for id-unrelated augmentation. Meanwhile, our proposed D-Mixup achieves better performance than the image-level Mixup and feature-level Mixup (F-Mixup) for id-related augmentation.

\begin{figure}
\centering
   \includegraphics[width=0.9\linewidth]{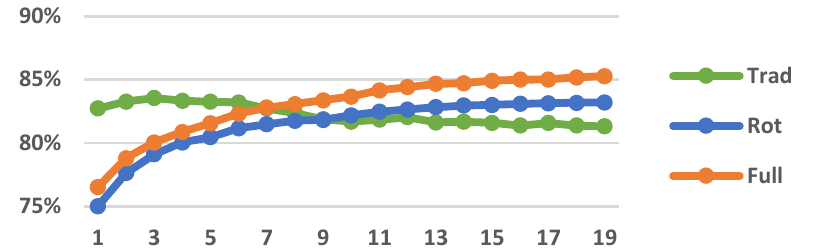}
   \caption{Normalized Mutual Information (NMI) during 20 joint training epochs on Market-1501. `Trad' refers to traditional data augmentation techniques. `Rot' refers to id-unrelated mesh-guided rotation. `Full' refers to combining id-unrelated mesh-guided rotation and id-related D-Mixup.}
\label{fig:NMI}
\end{figure}

\textbf{Effects on pseudo labels.} Robust identity representations should have a better intra-class compactness and inter-class separability, which leads to better pseudo label quality. We evaluate our pseudo label quality by measuring the Normalized Mutual Information (NMI)~\cite{Strehl2002ClusterE} between our pseudo labels and ground truth labels. As illustrated in Fig.~\ref{fig:NMI}, traditional data augmentation (Trad) works well at the beginning, but ends up in a worse quality. We argue that traditional data augmentation brings to the fore undesirable distortions on identity features, which easily leads to over-fitting for id-sensitive tasks. Deviating from that, GAN-based augmentation introduces more noise at the beginning, however avoids over-fitting in the final training epochs. In addition, our full GCL+ (Full) conducts both GAN-based id-unrelated and id-related augmentation, which achieves better pseudo label quality than only id-unrelated mesh-guided rotation (Rot).

\textbf{Contrastive loss.} To learn maximal invariance from generated image and memory stored image, we have formed three positive pairs for Rotation Contrast, namely $(f, f_{pos})$, $(f, f_{new}^{\prime})$ and $(f_{pos}, f_{new}^{\prime})$. By maximizing the similarity between these three positive pairs in Equation (\ref{view invariant loss}), (\ref{view invariant loss prime}) and (\ref{view invariant loss prime prime}), our objective is to build identity representations, which are invariant to instance-level pose, view-point and background variance. Meanwhile, we use identity prototypes and mixed prototypes in Mixup Contrast to learn a smoother class-level decision boundary with Equation (\ref{prototype contrast}) and (\ref{mixup contrast}). To confirm the contribution from these contrastive losses, we gradually add each into our framework and report the corresponding results in Table~\ref{table:vi vi2 vi3}. The results indicate that our proposed contrastive losses effectively contribute to learning robust representations for unsupervised person ReID.

\subsubsection{Comparison with state-of-the-art methods}
\textbf{Image-based person ReID.} We compare our proposed GCL+ with state-of-the-art unsupervised ReID methods under three purely unsupervised and four unsupervised domain adaptation evaluation protocols. We evaluate the performance of GCL+ with different baselines, including MMCL~\cite{Wang_2020_CVPR}, JVTC~\cite{li2020joint} and ACT~\cite{yang2020asymmetric}, to demonstrate the generalizability of our proposed method.

Under the fully unsupervised setting, we report associated results on Market-1501, DukeMTMC-reID and MSMT17 dataset in Table~\ref{table:fully unsupervised comparison}. We firstly provide results of state-of-the-art methods, including BUC~\cite{Lin2019ABC}, SoftSim~\cite{Lin2020UnsupervisedPR}, TSSL~\cite{wu2020tracklet}, MMCL~\cite{Wang_2020_CVPR}, JVTC~\cite{li2020joint}, JVTC+~\cite{li2020joint}, MetaCam~\cite{yang2021joint}, as well as our previous work GCL~\cite{Chen_2021_joint}, on the three datasets. Our proposed method GCL+ significantly improves the unsupervised person ReID performance from the three baselines MMCL, JVTC and JVTC+. The proposed new D-Mixup and Mixup Contrast in our framework GCL+ consistently surpasses the performance of our previous work GCL with the three different baselines. With the strong baseline JVTC+, our method achieves state-of-the-art performance on the three datasets. 

Under the unsupervised domain adaptation setting, we report related results on four mainstream benchmarks, including Duke$\to$Market, Market$\to$Duke, Market$\to$MSMT17 and Duke$\to$MSMT17 in Table~\ref{table:UDA comparison}. Our proposed method GCL+ additionally achieves better performance than state-of-the-art methods, including ECN \cite{zhong2019invariance}, PDA \cite{li2019cross}, CR-GAN \cite{chen2019instance}, SSG \cite{fu2019self}, MMCL \cite{Wang_2020_CVPR}, ACT \cite{yang2020asymmetric}, DG-Net++ \cite{Zou2020JointDA}, JVTC \cite{li2020joint}, ECN+ \cite{zhong2020learning}, JVTC+ \cite{li2020joint}, MMT \cite{ge2020mutual}, CAIL \cite{luo2020generalizing}, MetaCam \cite{yang2021joint}, as well as our previous work GCL~\cite{Chen_2021_joint}. Among these methods, PDA, CR-GAN and DG-Net++ share certain similarity with our proposed method GCL+, in that they are based on GAN. However, PDA and DG-Net++ used either 2D skeleton or random gray-scaled images as guidance, which could not preserve body shape information. Further, PDA, CR-GAN and DG-Net++ did not manipulate identity features to generate in-between identity images. CAIL \cite{luo2020generalizing} has considered cross-domain Mixup, where interpolated structures may introduce more noise on identity features. Our proposed D-Mixup does not suffer from such interpolated structures. In addition, cross-domain Mixup interpolates images from two domains, while our proposed D-Mixup interpolates intra-domain images, which is more flexible for fully unsupervised ReID.

\textbf{Video-based person ReID.} We compare our proposed GCL+ with state-of-the-art unsupervised video person ReID methods on MARS and DukeMTMC-VideoReID datasets. RACE~\cite{Ye2018RobustAE} and EUG~\cite{Wu2018ExploitTU} leverage a labeled video tracklet per identity to initialize their models. These one-example video-based ReID methods can not actually be considered as unsupervised. DAL~\cite{Chen2018DeepAL}, TAUDL~\cite{li2018unsupervised} and UTAL~\cite{li2019unsupervised} utilize camera labels of each tracklet and try to associate tracklets of a same person across different cameras. OIM~\cite{Xiao2017JointDA}, BUC~\cite{Lin2019ABC} and TSSL~\cite{wu2020tracklet} are fully unsupervised video person ReID methods. We use the fully unsupervised method BUC as our baseline. As shown in Table~\ref{table:unsupervised video reid comparison}, our proposed methods GCL (view-point augmentation) and GCL+ (view-point and in-between identity augmentation) significantly outperform previous unsupervised video-based person ReID methods.

\begin{table*}
\caption{Comparison with the state-of-the-art methods on two video-based re-ID datasets, MARS and DukeMTMC-VideoReID. The ``Labels" column indicates the labels used in each method. ``OneEx" denotes the one-example annotation per identity. ``Camera" refers to camera annotation. ``Baseline (BUC)" refers to our reproduced results.}
\label{table:unsupervised video reid comparison}
\centering
\scalebox{0.9}{
\begin{tabular}{l|c|cccc|cccc}
\hline
\multirow{2}{*}{Method} & \multirow{2}{*}{Labels}& \multicolumn{4}{|c}{MARS} & \multicolumn{4}{|c}{DukeMTMC-VideoReID} \\ \cline{3-10}
\multicolumn{1}{c|}{} &\multicolumn{1}{c|}{}& \multicolumn{1}{c}{mAP} & \multicolumn{1}{c}{R1}  & \multicolumn{1}{c}{R5} & \multicolumn{1}{c|}{R10}& \multicolumn{1}{c}{mAP} & \multicolumn{1}{c}{R1} & \multicolumn{1}{c}{R5}& \multicolumn{1}{c}{R10}\\ 
\hline

RACE~\cite{Ye2018RobustAE}&OneEx&24.5&43.2&57.1&62.1&-&-&-&-\\
EUG~\cite{Wu2018ExploitTU}&OneEx&42.4&62.6&74.9&-&63.2&72.7&84.1&-\\
DAL~\cite{Chen2018DeepAL}&Camera&23.0&49.3&65.9&72.2&-&-&-&-\\
TAUDL~\cite{li2018unsupervised}&Camera&29.1&43.8&59.9&72.8&-&-&-&-\\
UTAL~\cite{li2019unsupervised}&Camera&35.2&49.9&66.4&77.8&-&-&-&-\\
OIM~\cite{Xiao2017JointDA}&None&13.5&33.7&48.1&54.8&43.8&51.1&70.5&76.2\\
BUC~\cite{Lin2019ABC}&None&29.4&55.1&68.3&72.8&66.7&74.8&86.8&89.7\\
TSSL~\cite{wu2020tracklet}&None&30.5&56.3&-&-&64.6&73.9&-&-\\\hline
Baseline (BUC~\cite{Lin2019ABC})&None&32.0&51.1&66.5&71.6&67.1&72.9&86.2&90.0\\
GCL&None&48.6&64.8&77.5&82.0&75.9&80.1&90.5&93.7\\
GCL+&None&50.1&66.5&78.7&82.2&76.3&80.9&91.5&94.2\\
\hline
\end{tabular}}
\end{table*}

\subsection{Generation Quality Evaluation}

\subsubsection{Ablation study}
\begin{figure}
\centering
   \includegraphics[width=0.9\linewidth]{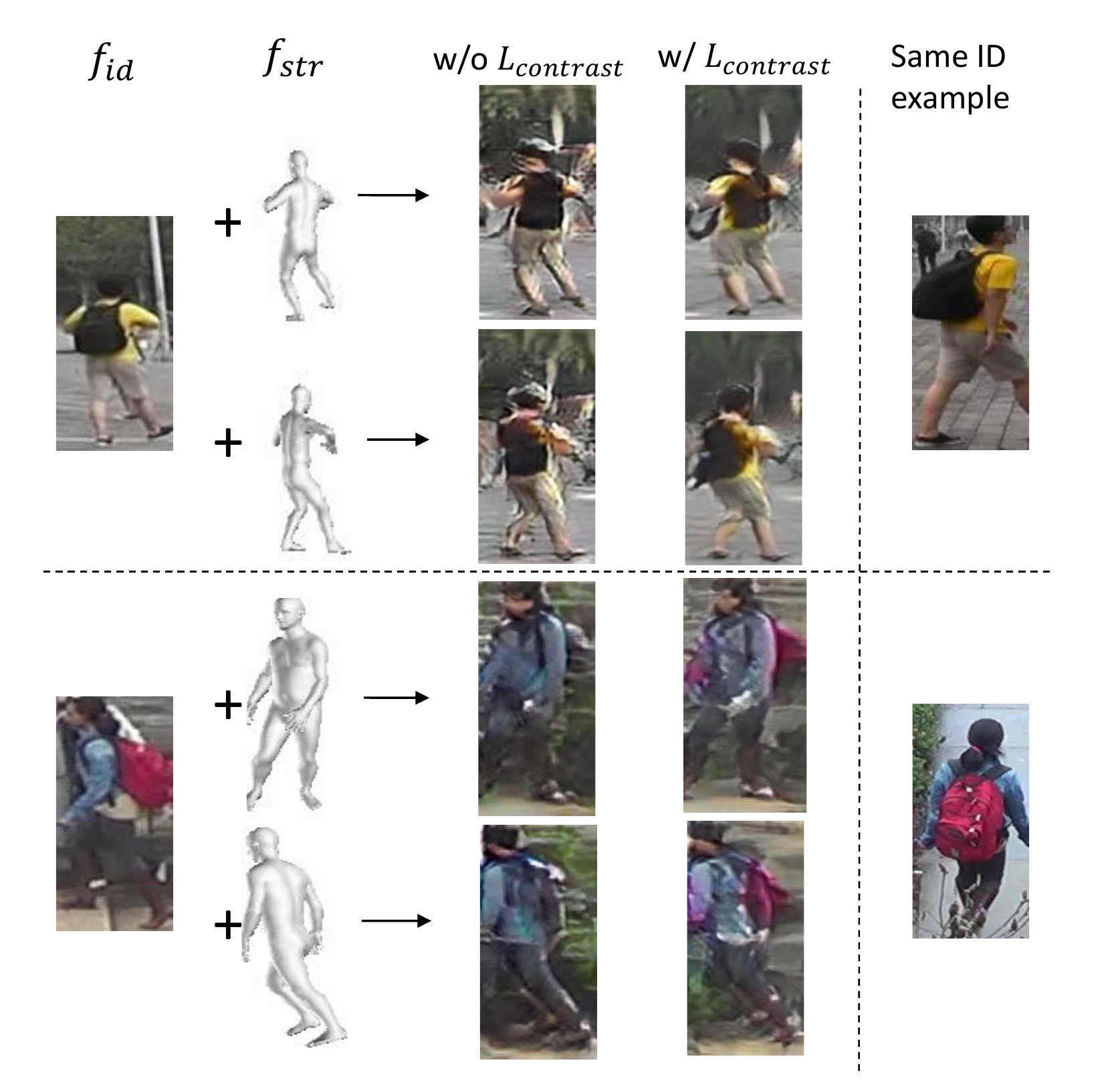}
   \caption{Qualitative ablation study on the effectiveness of contrastive loss in Eq.~(\ref{overall contrast loss}) for generation quality. $\mathcal{L}_{contrast}$ allows for preserving salient features from other views (yellow t-shirt for the first example and red backpack for the second example) in identity representations for generation in different poses and view-points.}
\label{fig:qualitative view-invariant}
\end{figure}

We conduct a qualitative ablation study, represented in Fig.~\ref{fig:qualitative view-invariant} to demonstrate that our proposed contrastive module can improve generative quality for person image generation. Unconditional GANs learn a data distribution via reconstruction and adversarial training of each image, which then generate new images that fit the learned distribution. However, unconditional GANs generate from features of a single image and neglect the shared features of different images of one person (or class). Conditional GANs generally use human-annotated identity labels to learn shared class-level features, which are more view-invariant. Our proposed GCL+ introduces an unsupervised way to learn view-invariant class-level features for person image generation by contrasting pseudo positive views. 

We illustrate two examples respectively from the Market-1501 and DukeMTMC-reID datasets in Fig.~\ref{fig:qualitative view-invariant} to validate the effectiveness of our proposed contrastive module for person image generation. Given a target person, a robust identity representation should contain salient features shared by the majority of observations in different view-points and poses. In the case that GCL+ is trained without $\mathcal{L}_{contrast}$, our generative module tends to focus only on salient features of original image (black backpack for the first example and blue jacket for the second example), while neglecting salient features of other images of the same person (yellow t-shirt for the first example and red backpack for the second example). The contrastive module ensures the consistency of identity features for generation in different poses and view-points.

\begin{figure*}
\centering
   \includegraphics[width=0.95\linewidth]{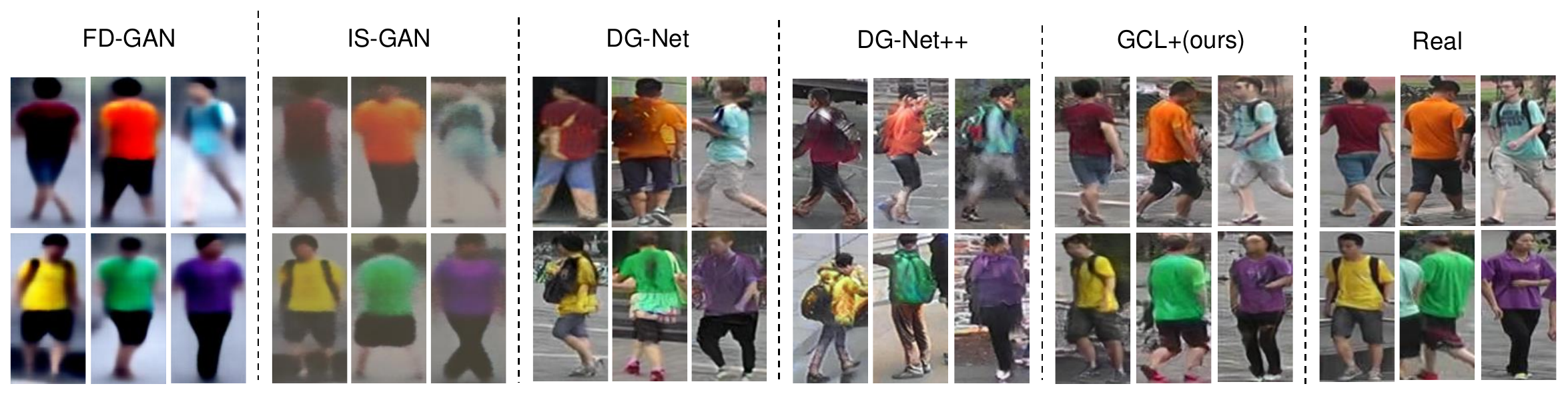}
\caption{Comparison of generated images on Market-1501 dataset. Examples of FD-GAN, IS-GAN, DG-Net, DG-Net++ and GCL+ are generated from same real images shown in the figure. We note that DG-Net++ and GCL+ are unsupervised methods.}
\label{fig:qualitative comparison}
\end{figure*}

\begin{table}
\caption{Examples of 3D mesh guided generation on DukeMTMC-reID dataset. }
\label{Tab: Rotation Duke}
\centering
\setlength\tabcolsep{2pt}
\begin{tabular}{ccccccccc}
  0\degree& &45\degree & 90\degree & 135\degree & 180\degree & 225\degree& 270\degree& 315\degree \\
\raisebox{-.4\height}{\includegraphics[scale=0.26]{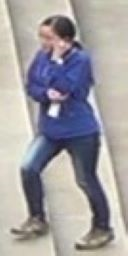}} &
$\to$&
\raisebox{-.4\height}{\includegraphics[scale=0.26]{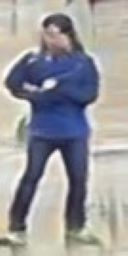}} & \raisebox{-.4\height}{\includegraphics[scale=0.26]{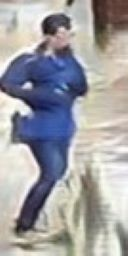}}& \raisebox{-.4\height}{\includegraphics[scale=0.26]{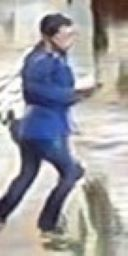}}& \raisebox{-.4\height}{\includegraphics[scale=0.26]{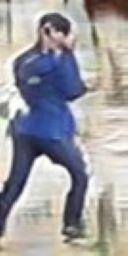}}& \raisebox{-.4\height}{\includegraphics[scale=0.26]{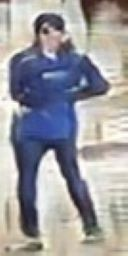}} &
\raisebox{-.4\height}{\includegraphics[scale=0.26]{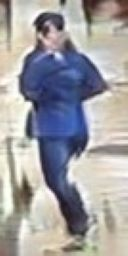}} &
\raisebox{-.4\height}{\includegraphics[scale=0.26]{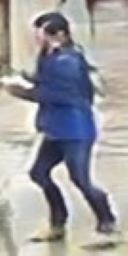}}
\\  
\raisebox{-.4\height}{\includegraphics[scale=0.26]{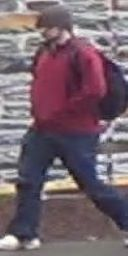}} &
$\to$&
\raisebox{-.4\height}{\includegraphics[scale=0.26]{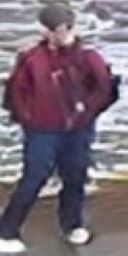}} & \raisebox{-.4\height}{\includegraphics[scale=0.26]{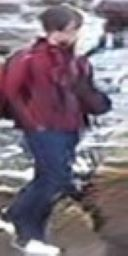}}& \raisebox{-.4\height}{\includegraphics[scale=0.26]{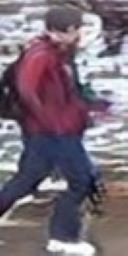}}& \raisebox{-.4\height}{\includegraphics[scale=0.26]{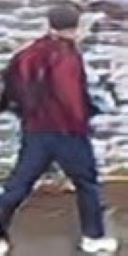}}& \raisebox{-.4\height}{\includegraphics[scale=0.26]{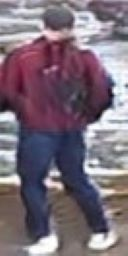}} &
\raisebox{-.4\height}{\includegraphics[scale=0.26]{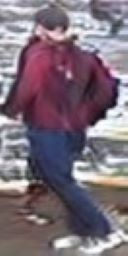}} &
\raisebox{-.4\height}{\includegraphics[scale=0.26]{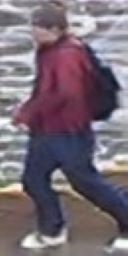}}
\end{tabular}
\end{table}

\begin{table}
\caption{Examples of 3D mesh guided generation on MSMT17 dataset. }
\label{Tab: Rotation MSMT}
\centering
\setlength\tabcolsep{2pt}
\begin{tabular}{ccccccccc}
  0\degree& &45\degree & 90\degree & 135\degree & 180\degree & 225\degree& 270\degree& 315\degree \\
\raisebox{-.4\height}{\includegraphics[scale=0.26]{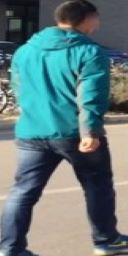}} &
$\to$&
\raisebox{-.4\height}{\includegraphics[scale=0.26]{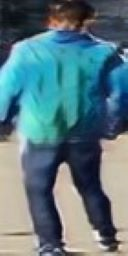}} & \raisebox{-.4\height}{\includegraphics[scale=0.26]{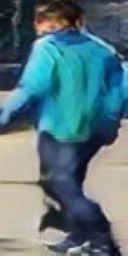}}& \raisebox{-.4\height}{\includegraphics[scale=0.26]{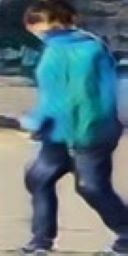}}& \raisebox{-.4\height}{\includegraphics[scale=0.26]{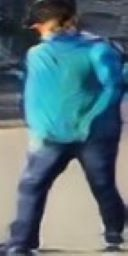}}& \raisebox{-.4\height}{\includegraphics[scale=0.26]{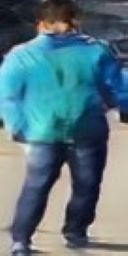}} &
\raisebox{-.4\height}{\includegraphics[scale=0.26]{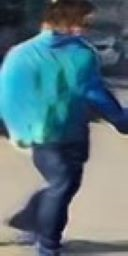}} &
\raisebox{-.4\height}{\includegraphics[scale=0.26]{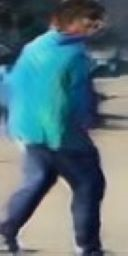}}
\\
\raisebox{-.4\height}{\includegraphics[scale=0.26]{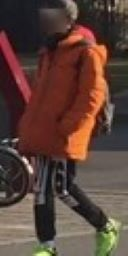}} &
$\to$&
\raisebox{-.4\height}{\includegraphics[scale=0.26]{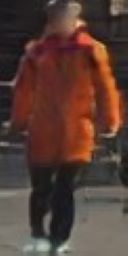}} & \raisebox{-.4\height}{\includegraphics[scale=0.26]{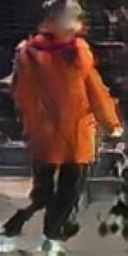}}& \raisebox{-.4\height}{\includegraphics[scale=0.26]{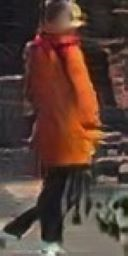}}& \raisebox{-.4\height}{\includegraphics[scale=0.26]{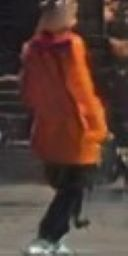}}& \raisebox{-.4\height}{\includegraphics[scale=0.26]{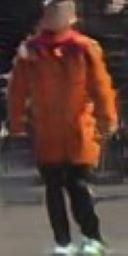}} &
\raisebox{-.4\height}{\includegraphics[scale=0.26]{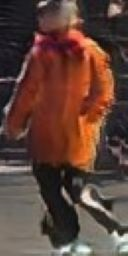}} &
\raisebox{-.4\height}{\includegraphics[scale=0.26]{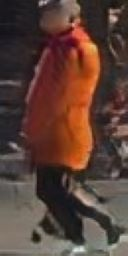}}\\
\end{tabular}
\end{table}

\subsubsection{Comparison with state-of-the-art methods}
\begin{figure}
\centering
\includegraphics[width=1\linewidth]{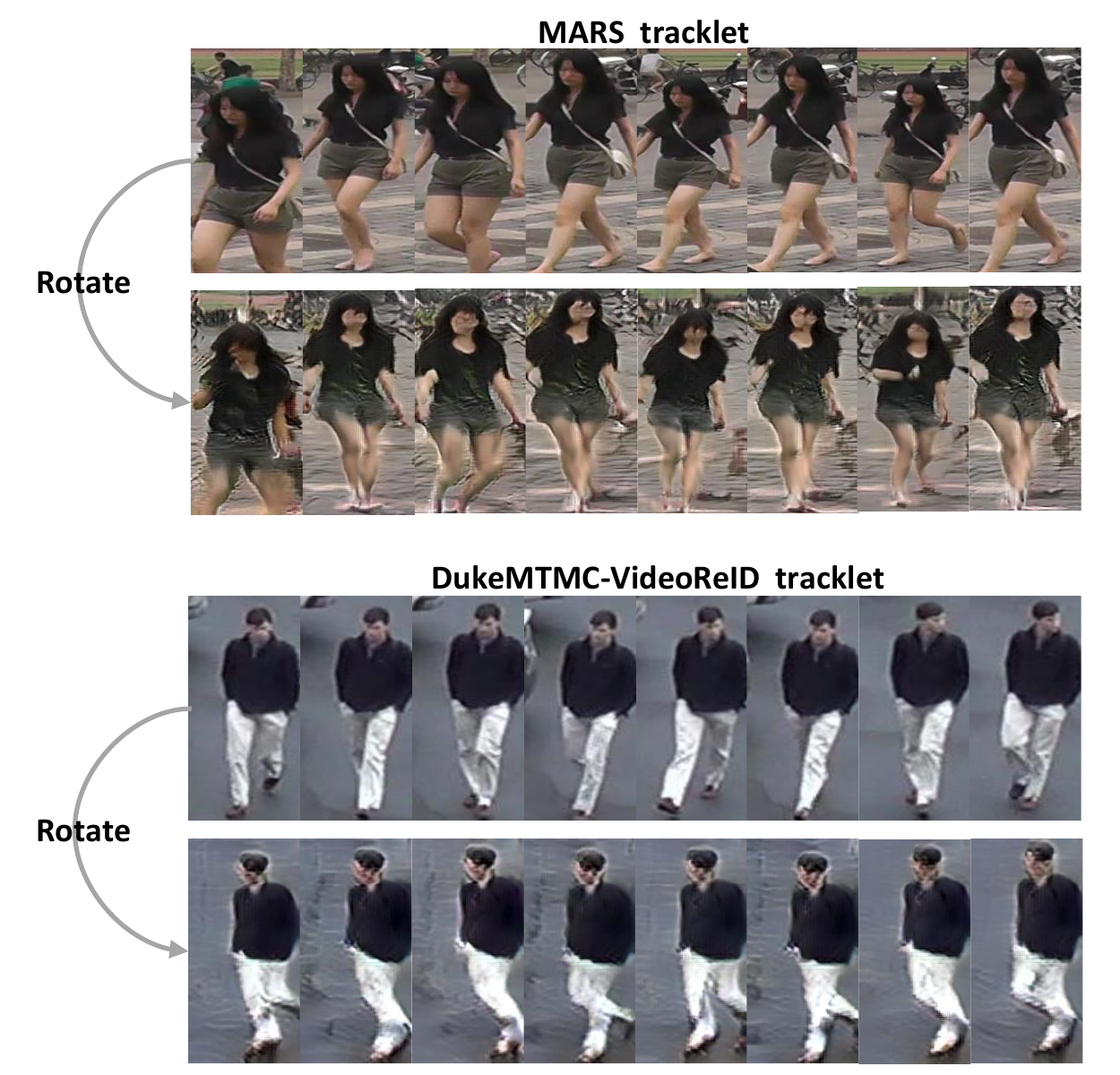}
\caption{Examples of tracklet frames before and after our view-point rotation. Tracklets are respectively sampled from MARS and DukeMTMC-VideoReID datasets.}
\label{fig_tracklet}
\end{figure}

We conduct a qualitative comparison between our proposed method GCL+ and state-of-the-art GAN-based person ReID methods, including FD-GAN~\cite{ge2018fd}, IS-GAN~\cite{NIPS2019_8771}, DG-NET~\cite{zheng2019joint} and DG-NET++~\cite{Zou2020JointDA}. We re-implement these GAN-based person ReID methods based on their published source code and generate six images per real image of the Market-1501 dataset, as shown in Fig.~\ref{fig:qualitative comparison}. FD-GAN, IS-GAN and DG-Net are supervised methods, which rely on human-annotated labels to learn robust identity-level features. We observe that images generated by FD-GAN and IS-GAN suffer from evident visual blur, which may lose detailed identity information after generation. Compared to FD-GAN and IS-GAN, DG-Net can generate sharper images. However, using randomly switched gray-scaled images as guidance is prone to result in incoherent body shape and carrying. More comparison on the generative quality between FD-GAN, IS-GAN, DG-Net and our method is provided in Supplementary Materials Section B. As an UDA method, DG-Net++ uses cross-domain gray-scaled images as guidance, which, however, shares same problems in generation as DG-Net. Different from DG-Net++, our proposed GCL+ is a fully unsupervised ReID method, which directly augments data diversity in the target domain without the need for a labeled source domain. Moreover, an image in GCL+ is generated from its own rotated mesh, which helps to conserve body shape information and does not add extra carrying structures. The generated images from GCL+ have higher quality and similarity to real images than other methods. To validate the generative quality on DukeMTMC-reID and MSMT17 datasets, we provide more examples in Table~\ref{Tab: Rotation Duke} and Table~\ref{Tab: Rotation MSMT}. Consistency in the id-related space and variance in the id-unrelated space validate the purity (disentanglement quality) of identity representations in our framework GCL+. We further provide tracklet examples before and after our view-point rotation for video-based person ReID in Fig.~\ref{fig_tracklet}. The results show that our method also works well for video-based person ReID.

\subsubsection{Failure case analysis}
\begin{figure}
\centering
\includegraphics[width=1\linewidth]{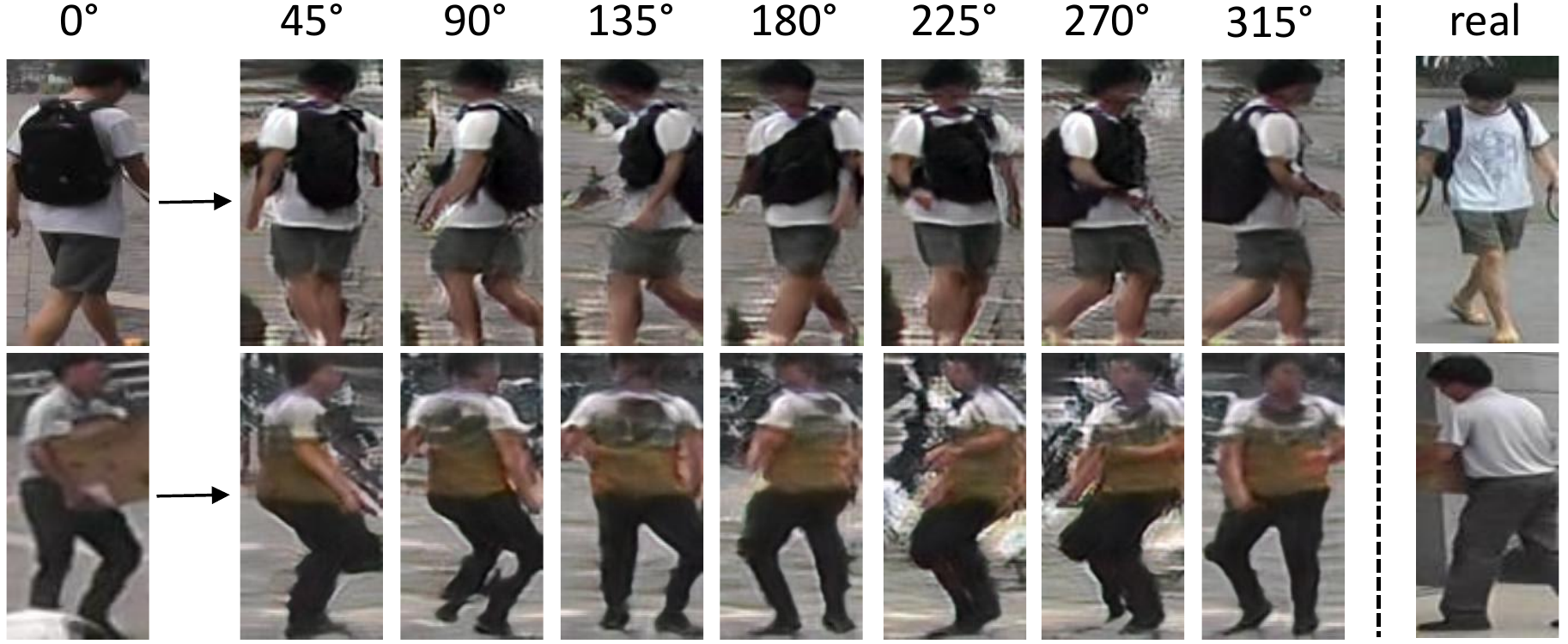}
\caption{Failure cases of rotation-based generation. First row: the backpack can be generated onto the front side. Second row: the carrying object can be generated onto the back side.}
\label{fig_failure_cases}
\end{figure}

We show some failure cases from the rotation generative model in Fig.~\ref{fig_failure_cases}. Actually, when there exists inconsistent front-side and back-side patterns, the rotation-based generation can hardly generate accurate images after large rotation. For example, the model may consider visual patterns only in the back side (backpack in the first row) and patterns only in the front side (carrying objects in the second row) as whole-body appearance features for generation. One possible solution is to use a 3D human-object arrangement mesh generator~\cite{zhang2020phosa} to help the generative model distinguish humans and objects.

\section{Conclusion}
In this paper, we propose an enhanced joint generative and contrastive learning (GCL+) framework for unsupervised person ReID. The framework is composed of a \textit{generative module} for data augmentation, as well as a \textit{contrastive module} aimed at learning invariance from generated variance. For the generative module, we propose a \textit{3D mesh guided GAN} to realize id-unrelated and id-related augmentation by respectively rotating 3D meshes as generation guidance and interpolating two identity representations. For the contrastive module, we design \textit{Rotation Contrast} and \textit{Mixup Contrast}, respectively for the two data augmentation techniques to learn robust identity representations. Extensive experiments are conducted to validate the superiority of the proposed GAN-based augmentation over traditional augmentation techniques for contrastive representation learning. The generative module benefits from learned robust identity representations that preserve fine-grained identity information for better generation quality. GCL+ outperforms state-of-the-art methods under both, fully unsupervised and unsupervised domain adaptation settings. Moreover, our contrastive module can be regarded as a contrastive discriminator in a GAN, which provides a new unsupervised approach for identity-preserving person image generation.



%



\ifCLASSOPTIONcompsoc
  \section*{Acknowledgments}
\else
  \section*{Acknowledgment}
\fi

This work has been supported by the French government, through the 3IA Côte d’Azur Investments in the Future project managed by the National Research Agency (ANR) with the reference number ANR-19-P3IA-0002.
The authors are grateful to the OPAL infrastructure from Université Côte d'Azur for providing resources and support.

\ifCLASSOPTIONcaptionsoff
  \newpage
\fi



%
\bibliographystyle{IEEEtran}
\bibliography{egbib}


%

\begin{IEEEbiography}[{\includegraphics[width=1in,height=1.25in,clip,keepaspectratio]{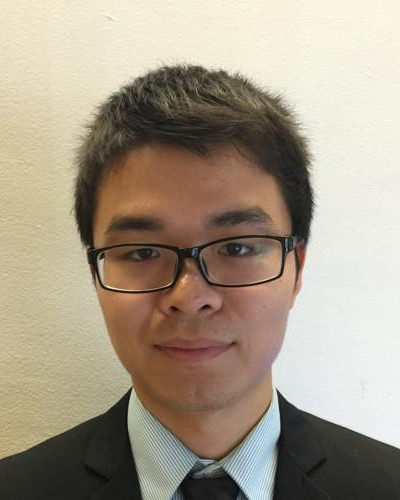}}]{Hao Chen}
received the B.S. degree from Wuhan University in 2014, and the M.S. degree from CentraleSupélec and Université Paris Saclay in 2017.
He is currently working towards his Ph.D. at Inria Sophia Antipolis and Université Côte d'Azur. His research
interests include person re-identification and unsupervised learning. Homepage: https://chenhao2345.github.io/.
\end{IEEEbiography}
\vspace{-1em}

\begin{IEEEbiography}[{\includegraphics[width=1in,height=1.25in,clip,keepaspectratio]{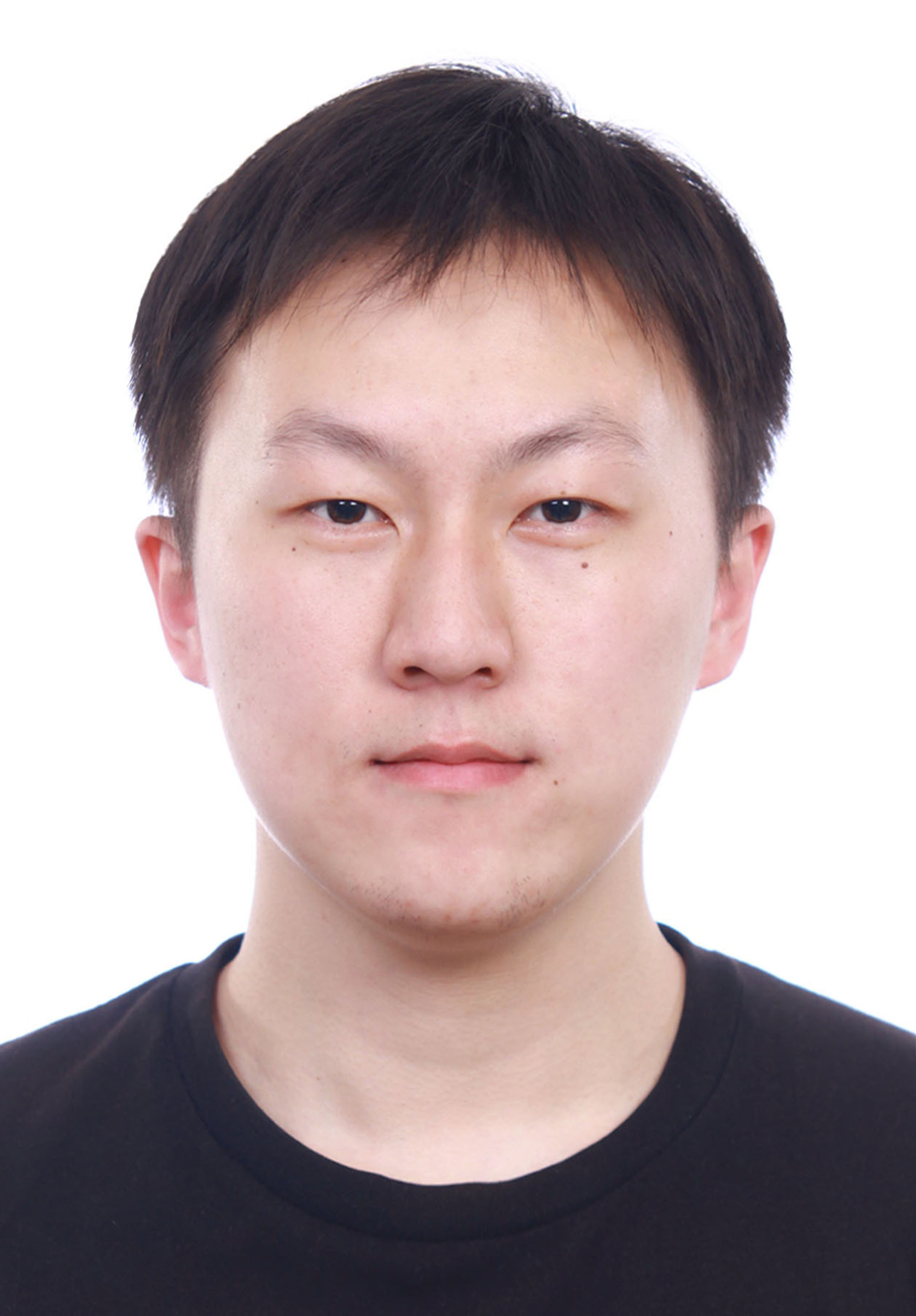}}]{Yaohui Wang}
received the B.S. degree from Xidian University in 2015, and the M.S. degree from ENSIIE and Université Paris Saclay in 2017.
He is currently working towards his Ph.D. at Inria Sophia Antipolis, STARS Team and Université Côte d'Azur. His current research focuses on image and video synthesis, activity recognition and representation learning.
\end{IEEEbiography}
\vspace{-1em}
\begin{IEEEbiography}[{\includegraphics[width=1in,height=1.25in,clip,keepaspectratio]{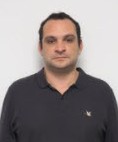}}]{Benoit Lagadec} is a Research Engineer at European Systems Integration. He currently works on developing video analysis solutions based on abnormal human behavior.  Previously, he worked in public research at Ifremer, where he was able to develop image processing algorithms adapted to the difficulty of underwater imaging : denoising, segmentation.
\end{IEEEbiography}
\vspace{-1em}
\begin{IEEEbiography}[{\includegraphics[width=1in,height=1.25in,clip,keepaspectratio]{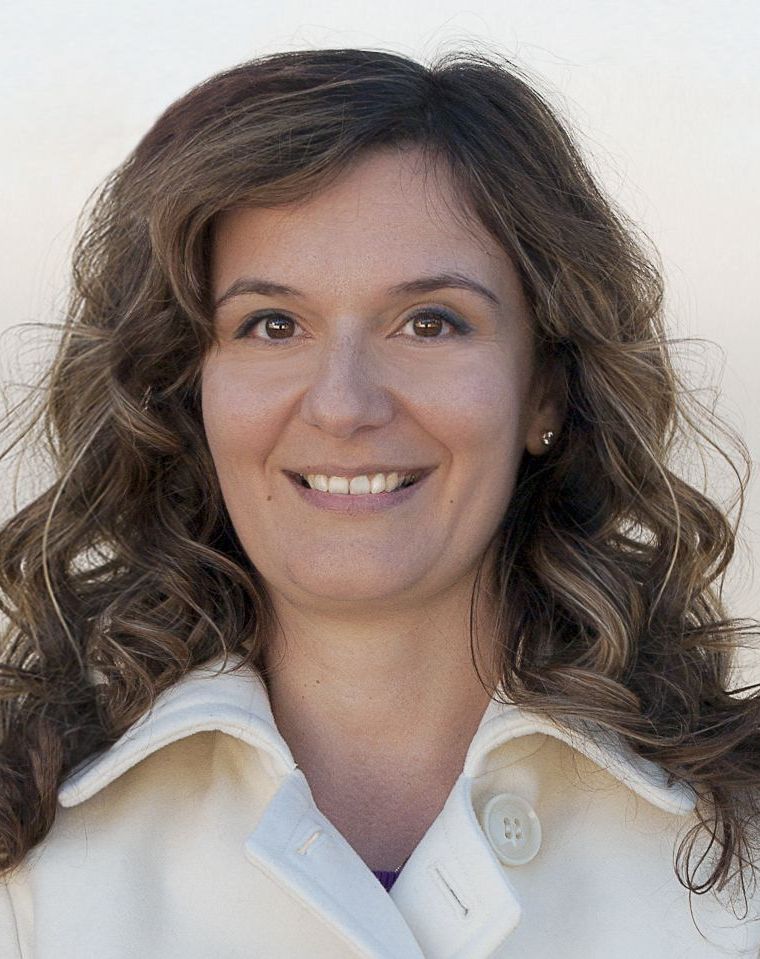}}]{Antitza Dantcheva}
is a Research Scientist
(CRCN) with the STARS team of INRIA Sophia
Antipolis, France. Previously, she was a Marie
Curie fellow at Inria and a Postdoctoral Fellow
at the Michigan State University and the West
Virginia University, USA. She received her Ph.D.
degree from Télécom ParisTech/Eurecom in image processing and biometrics in 2011. Her research
is in computer vision and specifically in
designing algorithms that seek to learn suitable
representations of the human face in interpretation
and generation. 
\end{IEEEbiography}
\vspace{-1em}
\begin{IEEEbiography}[{\includegraphics[width=1in,height=1.25in,clip,keepaspectratio]{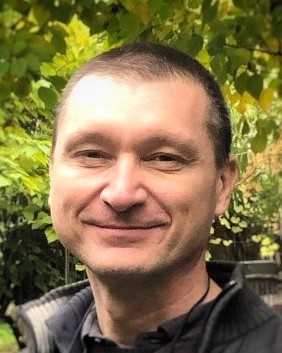}}]{Francois Bremond}
received the PhD degree from INRIA in video understanding in 1997, and he pursued his research work as a post doctorate at the University of Southern California (USC) on the interpretation of videos taken from Unmanned Airborne Vehicle (UAV). In 2007, he received the HDR degree (Habilitation a Diriger des Recherches) from Nice University on Scene Understanding. He created the STARS team on the 1st of January 2012. He is the research director at INRIA Sophia Antipolis, France. He has conducted research work in video understanding since 1993 at Sophia- Antipolis. He is author or co-author of more than 140 scientific papers published in international journals or conferences in video understanding. He is a handling editor for MVA and a reviewer for several international journals (CVIU, IJPRAI, IJHCS, PAMI, AIJ, Eurasip, JASP) and conferences (CVPR, ICCV, AVSS, VS, ICVS). He has (co-)supervised 26 PhD theses. He is an EC INFSO and French ANR Expert for reviewing projects.
\end{IEEEbiography}







\end{document}